%% file: cpm4sax.tex
\RequirePackage{fix-cm}
\documentclass [twocolumn,final] {svjour3} [15 Nov 2024]
\smartqed  

\AtBeginDocument{%
  \setlength{\oddsidemargin}{\dimexpr(\paperwidth-\textwidth)/2-1in}%
  \setlength{\evensidemargin}{\oddsidemargin}%
  \setlength{\topmargin}{%
    \dimexpr(\paperheight-\textheight)/2-\headheight-\headsep-1in}%
}

\usepackage{hyperref}
\usepackage{graphicx}
\usepackage{amsmath,amssymb}
\usepackage[shortlabels]{enumitem}
\usepackage{xcolor}
\usepackage[caption=false]{subfig}
\usepackage{float}
\usepackage{xcolor}
\usepackage{tabularx}
\usepackage{algorithm}
\usepackage{algpseudocode}
\usepackage{centernot}
\usepackage[normalem]{ulem}
\usepackage[T1]{fontenc}
\usepackage{lmodern}

\journalname{KÜNSTLICHE INTELLIGENZ Journal}

\usepackage[addedmarkup=bf,deletedmarkup=sout,authormarkup=none,commandnameprefix=always]{changes}
\definechangesauthor[color=red]{del}
\definechangesauthor[color=red]{add} 
\definechangesauthor[color=olive]{delr2}
\definechangesauthor[color=olive]{addr2} 

\begin{document}
%


\title{The WHY in Business Processes:\\ Discovery of Causal Execution Dependencies}

%
%
\author{Fabiana Fournier \and
Lior Limonad \and
Inna Skarbovsky \and 
Yuval David}
\authorrunning{F. Fournier et al.}
%
\institute{
IBM Research, Haifa, Israel\\
\email{\{fabiana,liorli,inna,yuval.david\}@il.ibm.com}}
\date{Received: 25 Oct 2023 / Accepted: 18 Nov 2024}
\maketitle              
%
\newif\ifshowcomments
\showcommentstrue
\ifshowcomments
    \newcommand{\mynote}[2]{\fbox{\bfseries\sffamily\scriptsize{#1}}{\small$\blacktriangleright$\textsf{#2}$\blacktriangleleft$}}
\else
    \newcommand{\mynote}[2]{}
\fi
\newcommand{\is}[1]{\textcolor{blue}{\mynote{Inna}{#1}}}
\newcommand{\lior}[1]{\textcolor{blue}{\mynote{LIOR}{#1}}}
\newcommand{\ff}[1]{\textcolor{purple}{\mynote{Fabiana}{#1}}}

\begin{abstract}
Unraveling the causal relationships among the execution of process activities is a crucial element in predicting the consequences of process interventions and making informed decisions regarding process improvements.
Process discovery algorithms exploit time precedence as their main source of model derivation. Hence, a causal view can supplement process discovery, being a new perspective in which relations reflect genuine cause-effect dependencies among the tasks. This calls for faithful new techniques to discover the causal execution dependencies among the tasks in the process.
To this end, our work offers a systematic approach to the unveiling of the causal business process by leveraging an existing causal discovery algorithm over activity timing.
In addition, this work delves into a set of 
conditions under which process mining discovery algorithms generate a model that is incongruent with the causal business process model, 
and shows how the latter model can be methodologically employed for a sound analysis of the process. Our methodology searches for such discrepancies between the two models in the context of three causal patterns, and derives a new view in which these inconsistencies are annotated over the mined process model. 
We demonstrate our methodology employing two open process mining algorithms, the IBM Process Mining tool, and the LiNGAM causal discovery technique. We apply it to a synthesized dataset and two open benchmark datasets.



\end{abstract}

\input{cpm4sax-intro}
\input{cpm4sax-background}
\input{cpm4sax-method}

\input{cpm4sax-exp-results}

\input{cpm4sax-related-work}

\input{cpm4sax-conclusions}

\begin{acknowledgements}
The research leading to these results received funding from the European Union’s Horizon research and innovation programme under grant agreements no 101094905 (AI4Gov), 101092639 (FAME), and 101092021 (AutoTwin).
\end{acknowledgements}

\bibliographystyle{splncs04}
\bibliography{mybibliography}




\end{document}

%% file: cpm4sax-intro.tex
\section{Introduction and Motivation}





Process mining (PM) aims to gain insights into an organization's business processes by analyzing event data recorded in its information systems, with the goal of improving business operations. 
Typical process mining techniques are inherently associational, approaching process discovery from a 
time precedence perspective, i.e., they discover ordering constraints among the process’ activities.
\chdeleted[id=del]{Causal inference techniques are largely concerned with discerning associative relationships and causal relationships.}
\chadded[id=add]{Such a}\chdeleted[id=del]{A}ssociative relationships are between two variables where a change in one variable is associated with a change in another. In contrast, causal relationships describe the connection between a cause and its effect, where the cause is an event that contributes to the production of another event, the effect~\cite{Pearl2000}. 

\chadded[id=add]{In this paper, we focus on employing causal discovery to identify a new type of process execution causal view.} Identification of causal relationships is key to the ability to reason about the consequences of interventions. One of the fundamental goals of causal analysis is not only to understand exactly what causes a specific effect but rather to be able to conclude if certain interventions account for the formation of certain outcomes, thus, being able to answer questions of the form: 
Does the execution of a certain activity in a loan approval process entail a delay in the handling of the application?
\chdeleted[id=del]{o}\chadded[id=add]{O}r if the same activity is skipped, may the process duration be shortened?





Typical process discovery techniques rely on time precedence among the activities in the event log, denoted by a \texttt{directly-follows} relationship \cite{van2016process}. This may sometimes be deceiving, due to the associative nature of such relationships, presenting an execution order that is different than the causal execution order inherent in the event log. Associative relationships are not enough to develop the causal understandings necessary to inform intervention recommendations as aforementioned. In order to understand which intervention should be made to improve a process outcome, we first must understand the causal chains that tie the unfolding of the processes generating our data.

\begin{figure*}[htbp]
    \centering
    {\includegraphics[scale = 0.22]{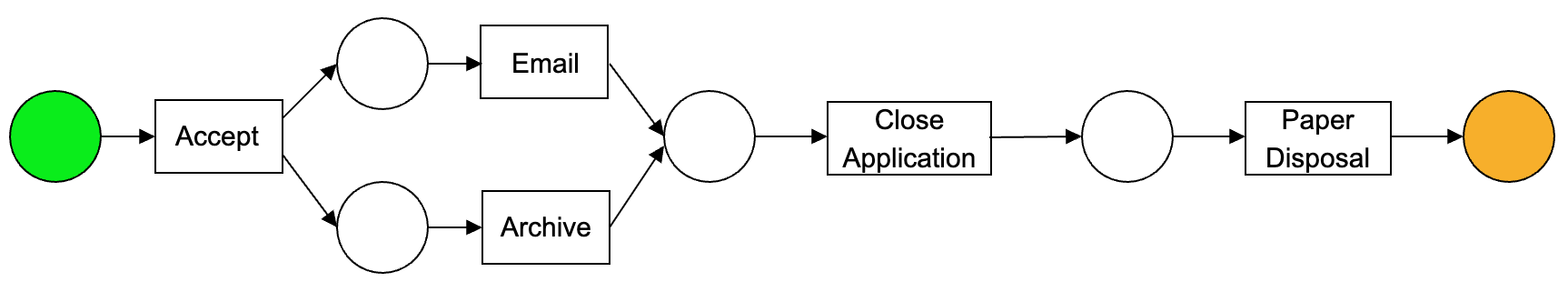}}
    \caption{The real business process.}
    \label{fig:bp}
\end{figure*}

As a simple example, we consider in Fig~\ref{fig:bp} a Petri-net~\cite{van2016process} depicting some of the final steps in a loan approval business process (BP), with its executions recorded in a log file. In this BP, the $Accept$ task precedes sending the client an email notification about the acceptance of the loan while also sending its records for archival retention. 
Note that the duration of the $Archive$ task is usually longer than that of the $Email$ task.
Subsequently, the application is closed in the system and all corresponding paperwork is disposed of.

\begin{figure*}[htbp]
\centering
\begin{tabular}{cc}
\parbox[c]{0.6\hsize}{\centering\includegraphics[scale = 0.22]{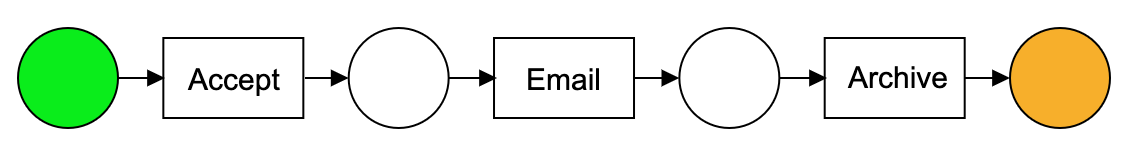}}
& 
\parbox[c]{0.4\hsize}{\centering\includegraphics[scale = 0.22]{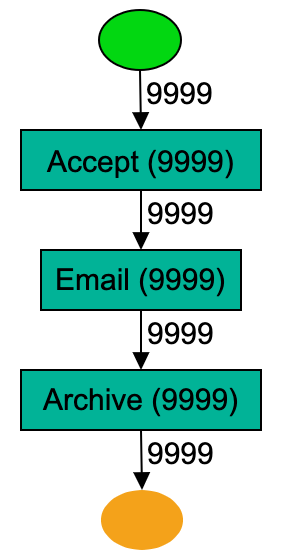}}     
\\
(a) $\alpha$-algorithm & (b) Heuristic algorithm
\\
\noalign{\vskip 1mm} 
\end{tabular}%
\caption{Process discovery results for the uniform case in the confounder pattern.}
\label{fig:uniform-case-pm-results}
\end{figure*}

Applying process discovery to the initial steps in this process (i.e., $Accept, Email, 
\text{and } Archive$)
can sometimes derive the graphs in Fig~\ref{fig:uniform-case-pm-results}. Our intuitive comprehension of such graphs attributes a causal meaning to their edges. That is, 
the links between the tasks in the graph are typically interpreted as
if the execution of the $Archive$ task depends on having the $Email$ task executed first. 
However, we know that in the real BP\chadded[id=add]{ (see Fig.~\ref{fig:bp})}, such dependency is not imposed, and may have arbitrarily emerged as a result of longer execution times for the $Archive$ task instances in the log. 

This may erroneously drive us to infer \chdeleted[id=del]{in the counterfactual case in which} \chadded[id=add]{that, for example, if} we do not send an email about a given loan acceptance, \chadded[id=add]{then, the loan}
\chdeleted[id=del]{we cannot deduce that it} is also not archived, or that if we \chadded[id=add]{shorten the execution time of} \chdeleted[id=del] {can expedite} the emailing task (e.g., by allocating more resources to it) then we can complete the archiving task earlier. 
\chdeleted[id=del]{Even in the cases when using more contemporary algorithms that utilize frequency thresholds (e.g., heuristic-miner and inductive-miner), yielding a result as depicted in Fig~\ref{fig:heuristic-sensitivity}(b) that retains parallelism among the $Email$ and $Archive$ tasks, the interpretation of such a graph structure is different.}
\chadded[id=add]{Even in the cases when using more contemporary algorithms that utilize frequency thresholds (e.g., heuristic-miner and inductive-miner), yielding a result that retains parallelism among the $Email$ and $Archive$ tasks in the real business process as depicted in Fig~\ref{fig:bp}, the interpretation of such a graph structure is different due to having a different meaning to the relations among the tasks as denoted by the Petri-net `place' notation. In a process view, an arrow denotes a time precedence relation. In a causal view, an arrow designates a causal execution dependence as defined in definition~\ref{def:def1} in section~\ref{sec:method}.}

\chdeleted[id=del]{While in a process view, such a structure}\chadded[id=add]{In our illustrative example, the process view} reflects individual time precedence dependencies between $Email$ and $Accept$ and between $Archive$ and $Accept$, with no \chdeleted[id=del]{dependence}\chadded[id=add]{precedence} between $Email$ and $Archive$ (i.e., parallel tasks)\chdeleted[id=del]{, a}\chadded[id=add]{. A} causal view of the same structure reflects the meaning of having the $Accept$ task being accountable for the execution of the two subsequent tasks\chadded[id=add]{, that is, the execution of $Email$ and $Archive$ depends on the prior execution of the $Accept$ task}.




We define a new type of a BP model in which all inter-task execution relations are causal as the Causal Business Process (CBP) model. \chadded[id=add]{We consider the CBP to be a new view about the BP since it assigns causal execution semantics to the fundamental relation among the tasks as later defined in Definition~\ref{def:def1}. Such a view aids process owners in gaining a sound understanding of the potential consequences of activity execution interventions (e.g., the effect of adding more resources to some activity).} Overall, we provide a novel method that draws from state-of-the-art causal discovery to reveal the CBP model from a given process event log based on the \texttt{execution times} of the activities. Specifically, we realize this by employing the LiNGAM causal discovery algorithm\footnote{\url{https://github.com/cdt15/lingam}}. The method is then also extended to tackle the cases where the \chadded[id=add]{discovered} BP model does not coincide with the CBP model, prescribing how to systematically annotate such discrepancies over the \chadded[id=add]{discovered} BP model. \chadded[id=add]{In our illustrative example this could be the case where the executions of the $Email$ and $Archive$ tasks are both dependent on the execution of the $Accept$ task (as shown in Fig~\ref{fig:lingam-results-confounder} and detailed henceforth), and the process discovery algorithms yield a result as in Fig~\ref{fig:uniform-case-pm-results}, where $Email$ directly-precedes $Archive$ and directly-follows $Accept$.} Our method is agnostic to the concrete process and causal discovery algorithms employed.




\chdeleted[id=del]{Acknowledging the importance of automated process adaptation and improvement as recently envisioned in the ABPMS manifesto~\cite{dumas2022augmented}, it is imperative that such improvements will rely on a genuine causal view of the process, regardless of the accidental appearance of correlations~\cite{pearl2009causality} (i.e., associative relationships) among its activity occurrences.}


We claim that the CBP model is a critical supplementary instrumentation underlying the ability to faithfully discover the 
logic that unfolds 
towards process execution outcomes, and consequently, improve and adapt underlying process executions.
\chadded[id=add]{Acknowledging the importance of automated process adaptation and improvement as recently envisioned in the ABPMS manifesto~\cite{dumas2022augmented}, it is imperative that such improvements also take into account a genuine causal execution view of the process, in addition to considering the temporal appearance of relationships within the process.}

\chadded[id=add]{In this paper we formulate a novel method for derivation of the causal execution view as a new view that overlays the discovered business process model. Subsequently, we demonstrate the applicability of our method on synthetic and real event logs. }

\chadded[id=add]{Our paper is structured as follows: Section~\ref{sec:background} presents relevant background about process discovery and causal discovery. Section~\ref{sec:method} describes our novel method for deriving CBPs and utilizing them to construct BP overlays that highlight inter-model discrepancies. General soundness and applicability assessments of the proposed method, followed by two pragmatic applications are presented in Section~\ref{sec:evaluation}. Respectively, the results of applying the method on one synthetic dataset and on two real-world applications for CBP to BP discrepancy analysis, and for inter-variant comparisons, are brought in Section~\ref{sec:results}. We review the related work on the intersection between causal discovery and business process discovery in the related work section (\ref{sec:related}). We wrap up the paper with a summary of conclusions and potential future work.}  

%% file: cpm4sax-background.tex
\section{Background}
\label{sec:background}

We use $\xrightarrow{c}$ to denote a causal execution relation. For example, $A \xrightarrow{c} B$ means that the execution of $A$ causes the execution of $B$. That is, without $A$ happening $B$ cannot happen, and given we know that $B$ occurred, we can deduce that $A$ had occurred \chadded[id=add]{(sometime)} before it. It is important to note that \chadded[id=add]{any} two \chadded[id=add]{time} correlated occurrences do not necessarily entail a causal execution relation. For example, let's assume that $Accept \xrightarrow{c} Email$, and also that $Accept \xrightarrow{c} Archive$. As a result, the occurrences of \chdeleted[id=del]{$Accept$,} $Email$ and $Archive$ will be correlated in the data. \chdeleted[id=del]{However, one cannot explain the archiving of a loan as being the result of emailing about its acceptance based solely on a correlation between the occurrences of the two events.}\chadded[id=add]{However, one cannot deduce a causal execution dependence between these two tasks, even if one always occurs immediately after the other.}
In our example, the correlation between $Email$ and $Archive$ is spurious, while the correlations between $Accept$ and $Email$ and between $Accept$ and $Archive$ reflect a causal relationship. 

\chdeleted[id=del]{Discovery of 
causal relationships from a given observational log may, in many cases, yield deceiving conditions in which dependencies could emerge or disappear due to unintended scoping of the observations on specific values or ranges of the variables observed, namely $conditioning$~\cite[p.~112]{pearl2018bookofwhy}.}
\chdeleted[id=del]{Some associations may be wrongly detected because of \chdeleted[id=del]{an} external, non-measured variables. Others may be omitted because of some inherent bias in the data that is recorded.} 

\subsection{Process Discovery}

Process discovery (PD) is one of the challenging tasks in PM, related to the effort of constructing a process model based on behavior that can be observed in a given event log \cite{van2016process,aalst2011process}.
The input for PD algorithms is an event log and the output is the business process model. 
Conventionally, an event log contains data related to a single process, where each $event$ refers to a single process instance, namely a $case$.  
An event log consists of several cases, i.e., multiple runs of a process \cite{van2016process}, where a case \chdeleted[id=del]{consists of}\chadded[id=add]{is a sequence of} events (at least one) \chadded[id=add]{referred to as a $trace$}. An event can also be related to some $activity$ or $task$. There exists a (time) order between events within the same case. Events include at least task name, timestamp, and case-id attributes.
Most fundamentally, PD algorithms rely on time precedence among the events in the log. 
For instance, if all log appearances (or a sufficiently high portion) of task $A$ occur before the ones of task $B$, PD may infer that task $A$ usually happens before task $B$, denoted by $A \rightarrow B$ in the inferred process model. However, such dependence does not necessarily imply that the occurrence of $A$ also \emph{causes} the occurrence of $B$. For example, it might be that a task $C$ is the real cause for both task occurrences.  



For the loan example assessment, we employed two PD representative algorithms, Alpha ($\alpha$) and Heuristic~\cite{van2016process,van2004workflow}. While $\alpha$ is strict, the Heuristic algorithm is more robust to noisy data, using a threshold in order to define if an edge between $A$ and $B$ exists, based on the number of times $A$ happened before $B$.
\chadded[id=addr2]{While we acknowledge that the $\alpha$ algorithm is somewhat outdated as it was the first algorithm developed for process mining, we consider its inclusion essential to demonstrate that the results of our algorithm, as presented in this paper, remain sound even when using the $\alpha$ algorithm as input, and how it coincides with the results of a more robust algorithm such as the heuristic miner. However, we limited the use of the $\alpha$ algorithm to synthetic datasets, where we could control the level of noise in the input. For real datasets, we replaced it with a variant of the contemporary inductive miner algorithm, which is embedded in the IBM Process Mining commercial product, as detailed in Sections~\ref{sec:evaluation} and~\ref{sec:results}.}
In both PD algorithms, there can be cases in which certain forms of \chdeleted[id=del]{causal} execution relations are misinterpreted from a causal point of view, \chdeleted[id=del]{and}\chadded[id=add]{when} modeled as a sequence of three tasks (Fig~\ref{fig:uniform-case-pm-results}).
This could occur either in logs where one task always occurs before another task, or where the order of occurrence between the two tasks reverses in a relatively small subset of cases.

\subsection{Causal Discovery}
Causal inference and causal discovery are the two main pillars of causal analysis. While causal discovery \chdeleted[id=del]{is responsible for}\chadded[id=add]{focuses on} analyzing and creating models that illustrate the relationships inherent in the data~\cite{peters2017elements,pearl2009causality,spirtes2000causation}, causal inference is the process of drawing a conclusion about a causal connection based on the conditions of the occurrence of an effect~\cite{YAO2021,cunningham2021causal,hernan2023causal}. In this work, we focus on causal discovery \chadded[id=add]{(CD)} \chdeleted[id=del]{and adapt for its employment} over process execution times\chdeleted[id=del]{,} as inherently recorded in \chadded[id=add]{the timestamps of the events in} process event logs\chdeleted[id=del]{, for uncovering of CBPs}.

More concretely, \chdeleted[id=del]{causal discovery}\chadded[id=add]{CD} aims at constructing causal graphs from data by exploring hypotheses about the causal structure~\cite{shimizu2022book}. Our focus is on 
the exploration of relations among task executions in a process for the sake of discovering the CBP.
Additional assumptions, such as functional forms and distributions, are often required to identify the causal graph from the data. In a typical \chdeleted[id=del]{setting of causal discovery}\chadded[id=add]{CD setting}, the causal graph is assumed to be a Directed Acyclic Graph (DAG), and all the common causes of observed variables are themselves observed \chadded[id=add]{(i.e., are present in the event log)}.


\chadded[id=add]{When intervention is possible}\chdeleted[id=del]{Concretely}, if $t(A)$ denotes the execution time for task $A$, and considering any two viable interventions $c_1$ and $c_2$ on this execution time \chadded[id=add]{(e.g., by adding more resources to it)}, one may conclude the execution time of any other task $B$ to be causally dependent on it if: 
\[p(t(B)\chadded[id=add]{=b}|do(t(A)=c_2)) \neq p(t(B)\chadded[id=add]{=b}|do(t(A)=c_1))\] where \chadded[id=add]{$p$ is a probability and} `do' designates an intervention on the execution time. \chadded[id=add]{There could be additional tasks (a.k.a confounders) that may be commonly affecting the duration of tasks $A$ and $B$. Hence, to determine the direct effect between these two tasks, all commonly affecting tasks should be handled (e.g., weighted in as covariates) to remove their effect.}
Subsequently, the elimination of all common causes affecting $t(A)$ and $t(B)$ allows for the removal of the `do' operator, and assessing from the process log directly whether~\cite{pearl1995}:  
\[p(t(B)\chadded[id=add]{=b}|t(A)=c_2) \neq p(t(B)\chadded[id=add]{=b}|t(A)=c_1)\]
Without any further assumptions about a given dataset, it may only be possible to distinguish between the independence of t(A) and t(B) and the dependence between the two activities. 

However, \chadded[id=add]{in observational circumstances as in our case, where intervention is not viable,} we may not be able to tell whether t(A) depends on t(B) or vice versa. \chadded[id=add]{Contemporary CD algorithms disentangle this issue of} \chdeleted[id=del]{I}\chadded[id=add]{i}dentifying the direction of dependence \chdeleted[id=del]{could be disentangled} with some additional assumptions about the dependence concerning its functional form and error distribution. 
Particularly, \chadded[id=add]{if}\chdeleted[id=del]{when} we can model the execution dependence between activities $A$ and $B$ as a linear relation\chadded[id=add]{ship, we can represent it} as follows:
\[ t(A) = e1 \]
\[ t(B) = \beta_{AB} \cdot t(A) + e2 \]
In a typical business process, given such settings, it is plausible to consider such a linear form of dependence between $t(A)$ and $t(B)$,  where the duration of both activities' executions may be captured by \chadded[id=add]{a non-Gaussian}\chdeleted[id=del]{some uniform or exponential} distribution \chdeleted[id=del]{(i.e., non-Gaussian)}\chadded[id=add]{(e.g., uniform or exponential)}. For simplicity, we also assume negligible transition times \chadded[id=add]{between activities} (i.e., $\beta_{AB}=1$)\chadded[id=add]{. This assumption can}\chdeleted[id=del]{, that could} be relaxed by capturing the transitions themselves as \chdeleted[id=del]{another activity}\chadded[id=add]{explicit activities}.

In this work, we employed the Linear Non-Gaussian Acyclic Model (LiNGAM) algorithm\chadded[id=add]{ for the discovery of the CBP}, given its main assumptions hold in the typical configuration of a process event log, and in particular adhere to  \chadded[id=add]{non-Gaussian} distribution of time as a continuous variable \chdeleted[id=del]{and its arbitrary variance being non-Gaussian}. These assumptions conform to the conventional modeling of business processes with activity durations drawn from \chdeleted[id=del]{uniform or exponential}\chadded[id=add]{non-Gaussian} distributions, and also that cycles in a typical business process are finite and can be eliminated with techniques such as k-loop unrolling.
Specifically, LiNGAM~\cite{shimizu2006linear} assumes that a causal model is composed of linear functions connecting the variables and that there are no unobserved confounders. Also, according to~\cite{hoyer2008nonlinear}, the linearity assumption can be in fact relaxed under additive noise assumption. 


\chdeleted[id=del]{LiNGAM gives a way to differentiate between $A \xrightarrow{c} B$ and $B \xrightarrow{c} A$ where $\xrightarrow{c}$ meaning causal relationship \cite{shimizu2006linear}, such that changing $B$ would not trigger a change in $A$ in the former, but a change in $A$ will trigger a change in $B$. This is achieved in LiNGAM by regressing both directions, i.e., in the bi-variate case estimate $f$ such that $y=\hat{f(x)}$, then deduce the residuals $u_x = \hat{f(x)} - y$ and check if $x \perp u_x$ is met. Similarly, do the same for $y$, i.e., $x = \hat{f(y)} , u_y = \hat{f(y)} - x$ and check again for independence of $y \perp u_y$. If both directions are independent, then no edge between $x$ and $y$ will be inferred. When dependence exists only in one direction, for example, $x \not\perp u_x$, then $y \xrightarrow{c} x$ is inferred. Conforming to its assumptions, there cannot be dependence in both directions.}
\chadded[id=add]{LiNGAM gives a way to differentiate between $x \xrightarrow{c} y$ and $y \xrightarrow{c} x$ where $\xrightarrow{c}$ meaning causal relationship \cite{shimizu2006linear}, such that changing $y$ would not trigger a change in $x$ in the former, but a change in $x$ will trigger a change in $y$. This is achieved in LiNGAM by regressing both directions, i.e., in the bi-variate case estimate $f$ such that $y=\hat{f(x)} + u_x$, then deduce the residuals $u_x = y - \hat{f(x)}$ and check if $x \perp u_x$ is met (i.e., $\perp$ designating independence). Similarly, do the same for $y$, i.e., $x = \hat{f(y)} + u_y , u_y = x - \hat{f(y)}$ and check again for independence of $y \perp u_y$. If both directions are independent, then no edge between $x$ and $y$ will be inferred. When dependence exists only in one direction, for example, $x \not\perp u_x$, then $y \xrightarrow{c} x$ is inferred. There cannot be dependence in both directions.}

\subsection{Causal Patterns}
\label{sec:causal-patterns}

\chadded[id=add]{The} \chdeleted[id=del]{D}\chadded[id=add]{d}{}iscovery of 
causal relationships from \chdeleted[id=del]{a given observational log may in many cases yield deceiving conditions in which dependencies could emerge or disappear due to unintended scoping of the observations on specific values or ranges of the variables observed, namely $conditioning$~\cite[p.~112]{pearl2018bookofwhy}.}
\chadded[id=add]{observational logs can often lead to misleading conditions in which dependencies may appear or disappear due to unintended scoping of observations to specific values or ranges of the observed variables. This phenomenon is known as $conditioning$, as discussed in~\cite[p.~112]{pearl2018bookofwhy}.}
Some \chdeleted[id=del]{associations}\chadded[id=add]{dependencies} may be wrongly detected because of external, non-measured variables (Simpson's paradox). Others may be omitted because of some inherent bias in the data selection (Berkson's paradox)~\cite{Simpson-Pearl-2014,simpsonp1982}. 

\begin{figure}[htbp]
\centering
\begin{tabular}{ccc}
\parbox[c]{0.2\columnwidth}{\centering\includegraphics[scale = 0.15]{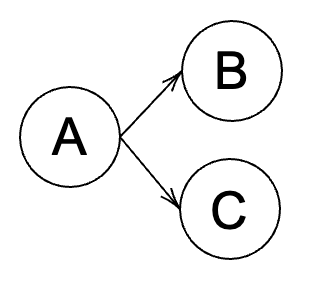}}
& 
\parbox[c]{0.4\columnwidth}{\centering\includegraphics[scale = 0.15]{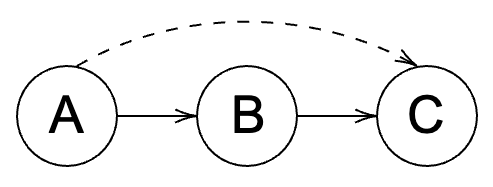}} 
&
\parbox[c]{0.2\columnwidth}{\centering\includegraphics[scale = 0.15]{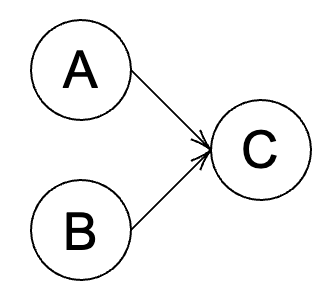}} 
\\
(a) Confounder & (b) Mediator & (c) Collider 
\\
\noalign{\vskip 1mm}  
\end{tabular}%
\caption{Causal patterns}
\label{fig:causal-patterns}
\end{figure}

As presented in \cite{pearl2018bookofwhy}, there are three causal relationship patterns or junctions that constitute the building blocks of any causal net structure (Fig~\ref{fig:causal-patterns}):

\begin{enumerate}[(a),topsep=3pt,leftmargin=*]
    \item 
    The confounder $A$ is a node that is a common cause for $B$ and for $C$ (the `affected' nodes). A confounder may create a correlation between $B$ and $C$ even if there is no direct causal relationship between them. 

    \item 
    The mediator $B$ is a node that transmits the effect of $A$ to $C$. If we condition on the mediator values we may not be able to observe the existence of a direct causal effect between $A$ and $C$. 

    \item 
    The collider $C$ is a node 
    that is caused by both $A$ and $B$. Conditioning on the collider $C$ may present a correlation between $A$ and $B$ that might be incorrectly interpreted as a causal relationship between the two. 
\end{enumerate}

Our aim is to identify the true inter-task causal relationships in the presence of any of these patterns in the event log, tackling the non-obvious effort to separate spurious correlation relations from causal relations. 
We assume \chdeleted[id=del]{execution traces of}\chadded[id=add]{all} the confounder tasks are included in the logs (i.e., known confounder \chadded[id=add]{assumption}). We show that state-of-the-art algorithms for process discovery may fail to genuinely capture the inter-task causal execution relationships, and thus assess the viability to employ the LiNGAM method \cite{shimizu2006linear} to annotate the result of the \chdeleted[id=del]{PM discovery}\chadded[id=add]{PD} algorithm according to the CBP as a new view.

%% file: cpm4sax-method.tex
\section{Our Method for Deriving Causal BP Overlays}
\label{sec:method}

We first formulate our definition of a causal execution dependence between tasks in a business process:

\begin{definition}[$A \xrightarrow{c} B$]
\label{def:def1}
A \texttt{causal execution dependence} implies that the time task $B$ executes is determined by the time task $A$ executes in a given process.
\end{definition}

While the dependence defined above is intrinsically a manifestation of some materialistic interaction between the inner workings of the two activities \cite{MILLER2019}, our definition is external, \chdeleted[id=del]{being focused}\chadded[id=add]{based} on the directed relationship (asymmetric) between the execution time property of any two tasks in the process. Such dependence implies that the execution time of task $B$ can only be determined based on the execution time of task $A$ and not vice versa. The relationship is directed such that $A \xrightarrow{c} B$ does not imply that $B \xrightarrow{c} A$. 
That is, changing the execution time of $A$ \chdeleted[id=del]{may affect}\chadded[id=add]{affects} the execution time of $B$, but not vice versa. 
\chdeleted[id=del]{Given an execution trace log for the two tasks,} 
\chadded[id=add]{In experimental settings, }altering \chdeleted[id=del]{between} different execution times for $A$ while \chdeleted[id=del]{fixating}\chadded[id=add]{fixing} the effects of all others potentially affecting variables (a.k.a., an `intervention') should present different distributions in the execution time of $B$ as a sign of \chadded[id=add]{a} causal execution dependence of $B$ on $A$. \chadded[id=add]{Practically, we employ LiNGAM to determine all causal execution dependencies from a given event log.} 
We distinguish between two possible modalities for the calculation of execution time, \textit{absolute} and \textit{relative}. The former relates to the time since process execution was initiated, while the latter relates to the time since the preceding activity finished its execution (see section~\ref{sec:pre-processing}).

There could be other dependencies among tasks, corresponding to other task attributes (e.g., number of allocated resources).
\chadded[id=add]{These other dependencies could be complementary to the CBP and can be denoted as follows: $A \xrightarrow[attribute]{c} B$.}
Such dependencies do not affect the conventional results of process mining, which are meant to cater to occurrence \chadded[id=add]{ordering}\chdeleted[id=del]{dependencies} among the tasks. 
In this work, our focus is strictly on the execution dependence that manifests itself via the timing of the tasks. \chdeleted[id=del]{There could be other extensions that could be seen as complementing the CBP with other inter-attribute causal dependencies, and where such dependencies could employ an extended notation such as $A \xrightarrow[attribute]{c} B$ (see sections \ref{sec:related} and \ref{sec:conclusions}).}

\begin{definition}[CBP]
\label{def:def2}
A \texttt{causal business process} is a process model in which inter-task relations among any of its tasks have a causal execution dependence. 
\end{definition}
It is important to stress that the \texttt{directly-follows} relation (i.e., ordering on time) is fundamentally different from the \texttt{causal} \texttt{execution} \texttt{dependence} in the sense that $A \xrightarrow{c} B \Rightarrow  A \rightarrow B$ but $A \rightarrow B \centernot\Rightarrow A \xrightarrow{c} B$. Furthermore, $A \rightarrow B \Rightarrow B \not\xrightarrow{c} A$, such that provided it is known that $B$ \texttt{directly-follows} $A$, the possible causal execution relation $B \xrightarrow{c} A$ is precluded. 
Thus, while the CBP view is complementary to the discovered BP view and cannot be inferred from it, we rely on its discovered \texttt{directly-follows} relations to narrow down the exploration space of the CBP. 

Contemporary tools for \chdeleted[id=del]{causal discovery}\chadded[id=add]{CD}, such as the LiNGAM algorithm, allow for a data-driven approach to reveal such causal execution dependencies under the aforementioned assumptions. 
Our particular focus in this work is on highlighting the discrepancies between the BP and the CBP with respect to the three causal patterns.
Concretely, with respect to definition \ref{def:def1} we further define the three patterns 
as follows:


\begin{definition}[$C \xleftarrow{c} A \xrightarrow{c} B$]
\label{def:def3-1}
A \texttt{confounding execution dependence} on $A$ is such that both the time task $B$ executes, and the time task $C$ executes, depend on the time the confounder task $A$ executes. 
\end{definition}


\begin{definition}[$A \xrightarrow{c} B \xrightarrow{c} C$]
\label{def:def3-2}
A \texttt{mediating execution dependence} on $B$ is such that the time task $B$ executes determines \chdeleted[id=del]{the extent to which}\chadded[id=add]{how much} the time task $C$ executes depends on the time task $A$ executes. 
\end{definition}


\begin{definition}[$A \xrightarrow{c} B \xleftarrow{c} C$]
\label{def:def3-3}
A \texttt{colliding execution dependence} on $B$ is such that the time the collider task $B$ executes depends on both the time task $A$ executes and the time task $C$ executes. 
\end{definition}

\textbf{Our hypothesis} 
is that given a \chadded[id=add]{timestamped} process log as an input, we can faithfully generate a CBP model with all causal execution dependencies and highlight its discrepancies compared to the process mining algorithm output.

As our validation shows, the highlighting of the inconsistencies can help with not only gaining fundamental insights about the process, but can also serve as a basis for better process improvement and interpretation of its execution outcomes.

\begin{figure}[htbp]
    \centering
    {\includegraphics[width=\columnwidth]{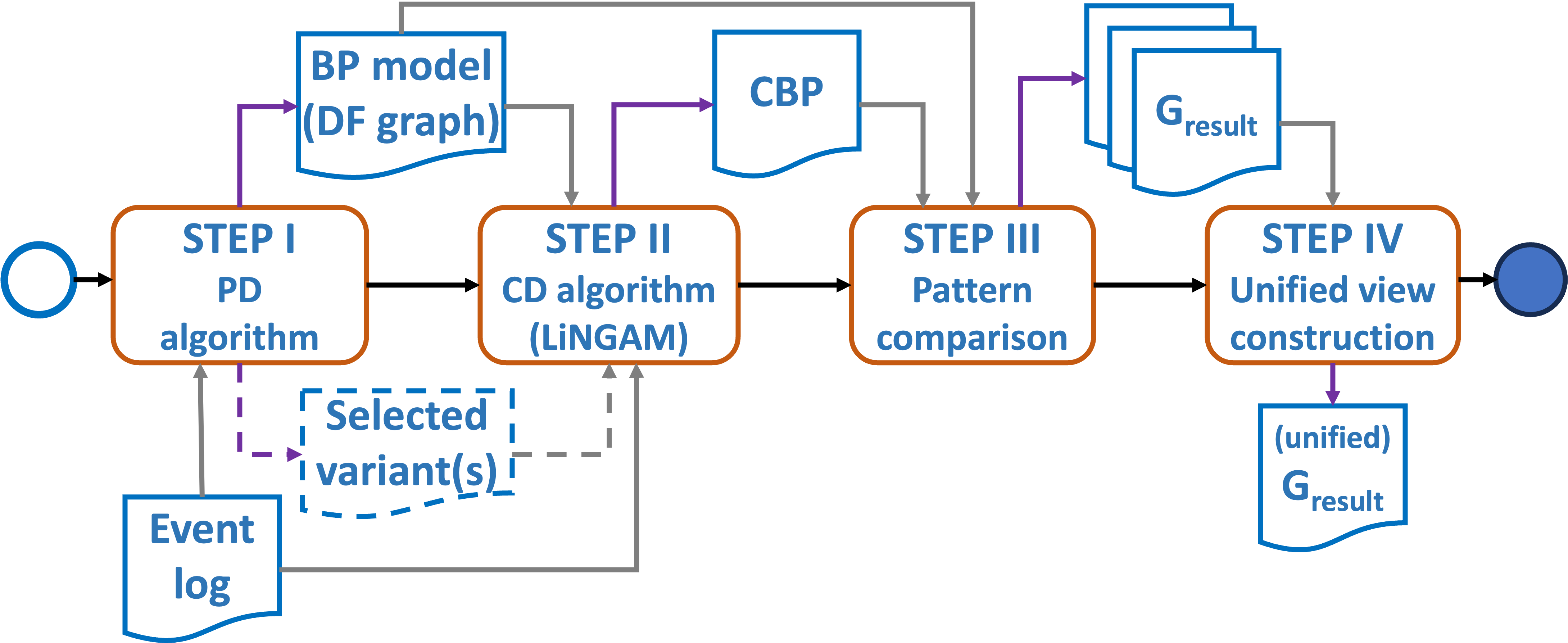}}
    \vspace*{-2mm}
    \caption{\chadded[id=add]{High-level illustration of our method.}}
    \label{fig:our-method}
\end{figure}

Our method to validate the above hypothesis caters to the following steps \chadded[id=add]{(see Figure~\ref{fig:our-method})}:
\begin{enumerate}[label=\Roman*.,topsep=2pt,leftmargin=*]
\item Apply \emph{any} state-of-the-art process mining algorithm for BP model discovery.
\item Apply the LiNGAM algorithm for CBP model discovery over the occurrence times of the activities.
\item Compare the CBP and the BP models with respect to the three patterns.
\item Construct a new view that combines the results in step 3 to highlight all the discrepancies.
\end{enumerate}

\chadded[id=add]{In the implementation of step I, we ensure that our method is decoupled from the selection of any specific process discovery technique. Our underlying assumption is that the discovered process model is correct, implying that our method is sound regardless of the particular PD algorithm used. Furthermore, the view constructed in step IV is relative to its input, regardless of its correctness.}

We note that in step II, our concrete choice of the LiNGAM algorithm may be relaxed to the use of other plausible \chdeleted[id=del]{causal discovery}\chadded[id=add]{CD} algorithms in the future, as long as the underlying assumptions of any such algorithm conform to the assumptions of conventional modeling of business processes (e.g., distribution of activity times).

To ensure the soundness of the resulting view with respect to the highlighting of causal execution dependencies, \chadded[id=add]{the} application of the above steps should conform to the following requirements:
\begin{itemize}[topsep=2pt,leftmargin=*]
    \item R1: For any case where a directly-follows relation $A \rightarrow B$ in the BP does \chdeleted[id=del]{n't}\chadded[id=add]{not} coincide with a causal execution dependence $A \xrightarrow{c} B$ in the CBP (i.e., the execution of $B$ is caused by some other activity $A'$ in the CBP)\chdeleted[id=del]{. Here}, the highlighting \chdeleted[id=del]{should both denote}\chadded[id=add]{denotes both} the directly-follows relation as \emph{not} being causal, and also \chdeleted[id=del]{denote} the relation $A' \xrightarrow{c} B$.
    \item R2: For any case where a directly-follows relation $A \rightarrow B$ in the BP coincides with a causal execution dependence $A \xrightarrow{c} B$ in the CBP, the highlighting denotes the directly-follows relation as being also a causal execution dependence.
\end{itemize}

\noindent
Next, we detail each of the steps in our method.




\subsection*{\textbf{Step I: Apply any state-of-the-art PD algorithm for BP model discovery}}

Process discovery is employed as a first step, to guide the overall discovery of causal execution dependencies. This is because while causal execution dependencies are not the same as temporal relations as \chdeleted[id=del]{aforementioned}\chadded[id=add]{mentioned above}, time-precedence is an essential condition (i.e., necessary but insufficient) in the occurrence of a causal one. That is, the absence of time-precedence also prohibits a causal execution dependence. As such, our method traverses over the discovered process model structure as a means to discover all causal execution dependencies. \chadded[id=add]{Thus, this step takes as an input the process event log and yields the discovered BP model in the form of a directly-follows graph as an output.}

The result of $PD$ caters to multiple process variants. A ``process variant'' refers to a specific sequence or path of activities in the process. Determining all process variants directly from an event log is essentially enumerating all unique traces in the log. For each group (trace), the sequence of activity names (event types) is extracted, and any duplicates (i.e., traces that represent the same sequence of activities) are identified and consolidated. This step yields all the unique traces, \chdeleted[id=del]{which are effectively}\chadded[id=add]{namely} the process variants. For various reasons, there can be many unique process variants. Some may be more \chdeleted[id=del]{highly} frequent in the log, while others may capture infrequent or even noisy behaviors. The \chadded[id=add]{CBP construction in the next step} is typically \chadded[id=add]{based on} \chdeleted[id=del]{constructed to capture} a unification of (user-selected) variants, while trading between variant \textit{completeness} and the model's ability to \textit{generalize} to cases that may have not yet been observed in the log. \chadded[id=add]{For flexibility of the method we keep the selection of particular (one or more) variants of interest optional, with the default of including all variants in the discovered BP.}\chdeleted[id=del]{For flexibility of the exploration, we keep the selection of particular (one or more) variants of interest external to the discovery process.} Such a variant selection is typically affected by the \chdeleted[id=del]{area in the PDM}\chadded[id=add]{segment of interest in the BP} subject to \chdeleted[id=del]{for the sake of intervening upon some of its observed states for the goal of process improvement}\chadded[id=add]{intervention}. Hence, we consider the remaining steps as being associated with the selection of a concrete set of one or more process variants \chadded[id=add]{as an additional output of step I}. Particularly, given a discovered process model \( M \) as the union of all variants in \( T \):
\[ M = \bigcup_{v \in T} v \] 
A single process variant \( v \) is one specific sequence (or list) of events from \( E \). Formally:
\[ v = (e_1, e_2, \ldots, e_n) \]
where \( e_i \in E \) for all \( i = 1, \ldots, n \), and \( n \) is the length of \chdeleted[id=del]{the} variant \( v \).

\subsection*{\textbf{Step II: Apply the LiNGAM algorithm for CBP model discovery over the occurrence times of the activities}}

The application of the LiNGAM algorithm involves two sub-steps. The first corresponds to the pre-processing of the data with respect to each pair of activities to be examined. The second sub-step prescribes a pair-wise order for LiNGAM invocations. \chadded[id=add]{It is important to note that a prerequisite to running LiNGAM is to test adherence to the non-Gaussian assumption. In general, the requirement in LiNGAM allows for a relaxation of this requirement to having one activity duration (at most) being drawn from a Normal distribution. In the (unlikely) case where more than one activity duration is normally distributed, the activity pair could be skipped in the second step.}

\subsection*{\textbf{Step II.a: Pre-processing}}
\label{sec:pre-processing}

Let \( E \) be the set of all event types, and \( C \) be the set of all cases. Let \( A \) be the set of all possible attributes. We define the initial input for the process as an event log \( L \), being a collection of tuples \( (e, c, t, N) \):
\[
L = \{(e_1, c_1, t_1, N_1), (e_2, c_1, t_2, N_2), (e_3, c_2, t_3, N_3), \ldots\}
\]

Where:
\begin{itemize}
    \item \( e \in E \) is an event type,
    \item \( c \in C \) is the case to which the event belongs,
    \item \( t \) is the timestamp indicating the time when the event occurred,
    \item \( N \) is a set of pairs \( (a, val) \) with \( a \in A \) and \( val \in \mathbb{R} \).
\end{itemize}


We transform the log \( L \) into a \chadded[id=add]{2-column table}\chdeleted[id=del]{tabular form} \( T \): 
\[ 
T = \{(c, [(e_1, t_1, N_1), (e_2, t_2, N_2), \ldots]) \}
\]

The tabular form \( T \) is a collection of rows, where each row \( r \in T \) corresponds to a unique case \( c \in C \), and \chdeleted[id=del]{each element in the row represents an event}\chadded[id=add]{an ordered list of events, ordered by their timestamps,} originally belonging to that case, where:
\begin{itemize}
    \item \( c \in C \) is the case ID,
    \item \( [(e_1, t_1, N_1), (e_2, t_2, N_2), \ldots] \) is \chdeleted[id=del]{a}\chadded[id=add]{an ordered} list of events, where \( e_1, e_2, \ldots \) are the event types, 
    \item \( t_1, t_2, \ldots \) are the timestamps of the events, and
    \item \( N_1, N_2, \ldots \) are the sets of name-value attribute pairs associated with each event.
\end{itemize}


We select all cases in $T$ in which their order of events matches the order of events in each variant of interest $v$.

Given:
 \( v = (e_1, \ldots , e_k) \) as an ordered set of events,
\begin{equation*}
\begin{split}
T^{v}=\text{Select}_{v} (T) = \{ & r \in T \mid r \text{ is of form } (c, s) \\
& s = [(e'_1, t'_1, N'_1), \ldots , (e'_k, t'_k, N'_k)] \\
& \text{such that } e'_j = e_j \\
& \text{for all } 1 \leq j \leq k \}
\end{split}
\end{equation*}

For any given set of variant $vs$ that is selected for the analysis, this step is repeated for each of the individual variants and then all sets are unified\chdeleted[id=del]{.}\chadded[id=add]{:}

\[ T^{vs} = \bigcup_{v \in vs} T^{v} \]



For any pair of events that is determined for comparison by the application of LiNGAM \chadded[id=add]{(see Step II.b)}, we extract a tuple of timestamps \chdeleted[id=del]{($ <e_i, e_j> $)}\chadded[id=add]{($ <t_i, t_j> $)}: 

\[
D = \{(t_i, t_j) \,|\, \forall c \in T^{vs}, \exists \, (e_i, t_i, N_i) \And (e_j, t_j, N_j) \, \in c\}.
\]

That is, we form a new set \( D \) consisting of tuples \( (t_i, t_j) \), where \( t_i \) and \( t_j \) are the timestamps of two extracted events in each case.


Next, we adjust the times extracted in each row in \( D \) with respect to some anchoring modality.
We refer to an \textit{anchor} as the point in time from which the activity execution duration is determined. Fundamentally, we consider one of two modalities:

\begin{enumerate}
    \item \textbf{Absolute} - Anchoring of all timestamps with respect to the initial execution time \( t_0 \) of each case. The adjustment can be represented as   
    \[
    D' = \{(t_i - t_0, t_j - t_0) \,|\, (t_i, t_j) \in D\}.
    \]
    \item \textbf{Relative} - Anchoring of the timestamps in each trace with respect to the timestamp of the event that precedes \( e_i \) in each case (i.e., \( t_{i-1} \)). The adjustment can be represented as
    \[
    D' = \{(t_i - t_{i-1}, t_j - t_{i-1}) \,|\, (t_i, t_j) \in D\}.
    \]
\end{enumerate}

Using $D'$ as an input, the adjusted timestamps are treated as variables and we apply the LiNGAM algorithm to detect causal relationships between $e_i$ and $e_j$ based on their corresponding adjusted timestamps $t_i$ and $t_j$.

\subsection*{\textbf{Step II.b: Pair-wise application of LiNGAM}}

With respect to any given set of variants $vs$ discovered by the PD algorithm, we apply the \chdeleted[id=del]{causal discovery}\chadded[id=add]{CD} algorithm (LiNGAM) in a pairwise manner over all ordered event pairs that preserve the original order in the set. 

That is, given a variant \( v \) represented as a sequence of events:
\[ v = (e_1, e_2, \ldots, e_n) \]
where \( n \) is the length of the variant, we denote the set of all ordered pairs in it as:
\[ P(v) = \{ (e_i, e_j) | 1 \leq i < j \leq n \} \]

The above is repeated for each $v \in vs$. We combine all event pairs into a single unified set and apply the LiNGAM algorithm to each pair of events in $P(vs)$:

\[ P(vs) = \bigcup_{v \in vs} P(v) \]

The application also includes a choice of anchoring \( \text{modality} \) which is either ``absolute'' or ``relative''. For a given set of variants to be examined, the same modality should be consistently employed across all LiNGAM applications. We then apply the LiNGAM algorithm to each individual element in \( P(vs) \) as follows:
\noindent
\[\text{setModality}(\text{``absolute'' or ``relative''}) \]
\[ \forall p \in P(vs):\]
\[\quad D' = \text{pre-process}(p,L) \quad \text{i.e., step IIa}\]
\[\quad G_p = \text{LiNGAM.apply}(D')\]
\noindent
Then, the union of all application results is:
\[ G_{L} = \bigcup_{p \in P(vs)} G_p \]

Where $G_{L}$ is the graph that denotes the discovered causal business process (CBP) that corresponds to the set of variants $vs$ in the \chdeleted[id=del]{$PD$}\chadded[id=add]{discovered BP}.

\chadded[id=add]{In our illustrative example, the output of step II.b is the graph $G_L:(V_L,E_L)$:
\[G_L = \bigcup(\dots \]
\[G_p:(V_p:\{Accept, Email\},E_p:\{(Accept, Email)\}),\]
\[G_p:(V_p:\{Accept,Archive\},E_p:\{(Accept,Archive)\}),\]
\[G_p:(V_p:\{Email,Archive\},E_p:\{\})\dots) =\]
\[(V_L:\{Accept, Email, Archive\},\]
\[E_L:\{(Accept, Email),(Accept,Archive)\})\]
This reflects a $G_L$ result in which the $Accept$ task is a confounder as denoted in Fig.~\ref{fig:lingam-results-confounder}.}

\subsection*{\textbf{Step III: Compare the CBP and the \chdeleted[id=del]{PDM}\chadded[id=add]{discovered BP models} with respect to the three patterns}}

Compare the discovered \chdeleted[id=del]{process model (PDM)}\chadded[id=add]{BP model} and the CBP models with respect to the three causal patterns: confounder, collider, and mediator. The comparison highlights the differences in terms of marking the \texttt{directly-follows} relations in the \chdeleted[id=del]{PDM}\chadded[id=add]{discovered BP} as either causal or non-causal, and also adding additional causal execution dependencies missing in the \chdeleted[id=del]{PDM}\chadded[id=add]{discovered BP}. All comparison results are merged into a new process perspective graph ($G_{result}$) that constitutes a causal discrepancy view.

\chadded[id=add]{Algorithm~\ref{alg:sequential-structure-1} applies to the case where the LiNGAM output DAG shows a confounder structure (i.e., being $A$ the confounder) while the PD algorithm discovers this structure as a sequence of $A \rightarrow B \rightarrow C$.}
In accordance with the aforementioned requirements, the generation of a new view that highlights the discrepancies between the two models corresponding to the confounder pattern should conform to: R1 - highlighting the relationship $B \rightarrow C$ as not being causal, and adding $A \xrightarrow{c} C$, and R2 - highlighting the relationship $A \rightarrow B$ as also being causal. We formally elaborate the realization of step \#3 in our method (\chadded[id=add]{see} sec\chadded[id=add]{tion} \ref{sec:method}) for the confounder pattern in algorithm~\ref{alg:sequential-structure-1}.

\algnewcommand\algorithmicforeach{\textbf{for each}}
\algdef{S}[FOR]{ForEach}[1]{\algorithmicforeach\ #1\ \algorithmicdo}

\begin{algorithm}
\caption{Given a \chdeleted[id=del]{PDM}\chadded[id=add]{discovered BP} and a corresponding CBP that may consist of a CONFOUNDER pattern}
\label{alg:sequential-structure-1}
\begin{algorithmic}[1] 
\Require 
    \State $G_{BP},G_L$ \Comment{Given BP and CBP graphs}
    \State $\forall t \in G_{BP} \implies \exists t' \in G_L$ \Comment{$t$ and $t'$ denote the same task in the respective models}
    \State $G_{result}\gets G_{BP}$ \Comment{$G_{result}$ highlights the $\Delta(G_{BP}-G_L)$}
    \ForAll {$(a,b,c) \in G_{BP}:a \twoheadrightarrow b, b \rightarrow c, a \not\rightarrow c$}
        \If{$\exists(a',b',c') \in G_L: a' \xrightarrow{c} b', a' \xrightarrow{c} c'$} \Comment{Check $a'$ is a confounder in $G_L$:}
            \If{$a \rightarrow b$} \Comment{Given $b$ directly-follows $a$}
                \State $G_{result}\gets$ \Call{$G_{BP}$.label}{$a \rightarrow b$, `C'}
            \Else \Comment{Given $b$ eventually-follows $a$}
                \State $G_{result}\gets$ \Call{$G_{BP}$.add}{$a \xrightarrow{c} b$} \Comment{added only if not already included}
            \EndIf
            \State $G_{result}\gets$ \Call{$G_{BP}$.add}{$a \xrightarrow{c} c$}
            \If{$b' \not\xrightarrow{c} c'$} \Comment{Check if $b'$ does not cause $c'$}
                \State $G_{result}\gets$ \Call{$G_{BP}$.label}{$b \rightarrow c$, `not C'}
            \EndIf
        \EndIf
    \EndFor
    \State \textbf{return} $G_{result}$
\end{algorithmic}
\end{algorithm}

\chdeleted[id=del]{Algorithm~\ref{alg:sequential-structure-1} applies to the case where the LiNGAM output DAG shows a confounder structure (i.e., being $A$ the confounder) while the PD algorithm discovers this structure as a sequence of $A \rightarrow B \rightarrow C$.} We note that we also check for $A \not\rightarrow C$ since such a sequence gives rise to the possibility of $B$ not being executed at all \chadded[id=add]{in the discovered BP. Provided $A$ is a confounder in this case, a variant in which only $C$ executes after $A$ is not feasible}.
In this case, the algorithm sets the resulting graph $G_{result}$ to initially match the sequential structure in the PD, adds the missing causal edge $A \xrightarrow{c} C$, and explicitly annotates the directly-follows edge $A \rightarrow B$ with a `$c$' symbol to indicate its essence of being causal. Lastly, if $B$ is also not a cause for the execution of $C$, the directly-follows relationship $B \rightarrow C$ is also denoted as being non-causal.

\chdeleted[id=delr2]{Algorithm~\ref{alg:sequential-structure-2} applies to the case where the LiNGAM output DAG shows a collider structure (i.e., being $C$ the collider) while the PD algorithm discovers this structure as a sequence of $A \rightarrow B \rightarrow C$.}
\chadded[id=addr2]{Algorithm~\ref{alg:sequential-structure-2} relates to a collider structure (i.e., being $C$ the collider) in the LiNGAM output DAG where the PD algorithm is a sequence $A \rightarrow B \rightarrow C$ as in Algorithm~\ref{alg:sequential-structure-1}}.
\chdeleted[id=delr2]{In accordance with the aforementioned requirements, the generation of a new view that highlights the discrepancies between the two models corresponding to the collider pattern should conform to: R1 - highlighting the relationship $A \rightarrow B$ as not being causal, and adding $A \xrightarrow{c} C$, and R2 - highlighting the relationship $B \rightarrow C$ as also being causal. We formally elaborate the realization of step \#3 in our method (sec \ref{sec:method}) for the collider pattern in algorithm~\ref{alg:sequential-structure-2}.}
\chadded[id=addr2]{Similarly, for the collider pattern, the generation of a new view to highlight the discrepancies between the two models should follow: R1 - marking the relationship $A \rightarrow B$ as non-causal and adding $A \xrightarrow{c} C$, and R2 - highlighting that the relationship $B \rightarrow C$ is also causal. The implementation of step \#3 in our method (sec \ref{sec:method}) for the collider pattern is formally described in algorithm~\ref{alg:sequential-structure-2}.}

\begin{algorithm}
\caption{Given a \chdeleted[id=del]{PDM}\chadded[id=add]{discovered BP} and a corresponding CBP that may consist of a COLLIDER pattern}
\label{alg:sequential-structure-2}
\begin{algorithmic}[1] 
\Require 
    \State $G_{BP},G_L$ \Comment{Given BP and CBP graphs}
    \State $\forall t \in G_{BP} \implies \exists t' \in G_L$ \Comment{$t$ and $t'$ denote the same task in the respective models}
    \State $G_{result}\gets G_{BP}$ \Comment{$G_{result}$ highlights the $\Delta(G_{BP}-G_L)$}
    \ForAll {$(a,b,c) \in G_{BP}:a \rightarrow b, b \twoheadrightarrow c, a \not\rightarrow c$}
        \If{$\exists(a',b',c') \in G_L: a' \xrightarrow{c} c', b' \xrightarrow{c} c'$} \Comment{Check $c'$ is a collider in $G_L$:}
            \If{$b \rightarrow c$} \Comment{Given $c$ directly-follows $b$}
                \State $G_{result}\gets$ \Call{$G_{BP}$.label}{$b \rightarrow c$, `C'}
            \Else \Comment{Given $c$ eventually-follows $b$}
                \State $G_{result}\gets$ \Call{$G_{BP}$.add}{$b \xrightarrow{c} c$} \Comment{added only if not already included}
            \EndIf
            \State $G_{result}\gets$ \Call{$G_{BP}$.add}{$a \xrightarrow{c} c$}
            \If{$a' \not\xrightarrow{c} b'$} \Comment{Check if $a'$ does not cause $b'$}
                \State $G_{result}\gets$ \Call{$G_{BP}$.label}{$a \rightarrow b$, `not C'}
            \EndIf
        \EndIf
    \EndFor
    \State \textbf{return} $G_{result}$
\end{algorithmic}
\end{algorithm}

\chdeleted[id=del]{Algorithm~\ref{alg:sequential-structure-2} applies to the case where the LiNGAM output DAG shows a collider structure (i.e., being $C$ the collider) while the PD algorithm discovers this structure as a sequence of $A \rightarrow B \rightarrow C$.} 
\chdeleted[id=delr2]{We note that we also check for $A \not\rightarrow C$ since such a sequence gives rise to the possibility of $B$ not being executed at all in the discovered BP. Provided $C$ is a collider in this case, a variant in which only $A$ executes before $C$ is not feasible.
In this case, the algorithm sets the resulting graph $G_{result}$ to initially match the sequential structure in the PD, adds the missing causal edge $A \xrightarrow{c} C$, and explicitly annotates the directly-follows edge $B \rightarrow C$ with a `$c$' symbol to indicate its essence of being causal. Lastly, if $A$ is also not a cause for the execution of $B$, the directly-follows relationship $A \rightarrow B$ is also denoted as being non-causal.}
\chadded[id=addr2]{Similarly, we check for $A \not\rightarrow C$ in the collider scenario, as this sequence suggests the possibility of $B$ being skipped in the discovered BP. If $C$ acts as a collider in this case, a variant in which only $A$ occurs before $C$ is deemed unfeasible. Here, the algorithm adjusts the graph $G_{result}$ to initially align with the sequential pattern in the PD, includes the missing causal edge $A \xrightarrow{c} C$, and labels the directly-follows edge $B \rightarrow C$ with a `$c$' to reflect its causal significance. Finally, if $A$ is not a causal factor for $B$’s execution, the directly-follows relationship $A \rightarrow B$ is also marked as non-causal.}

\chdeleted[id=delr2]{Algorithm~\ref{alg:sequential-structure-3} applies to the case where the LiNGAM output DAG shows a mediator structure (i.e., being $M$ the mediator) while the PD algorithm discovers this structure as a sequence of $A \rightarrow M \rightarrow B$.
In accordance with the aforementioned requirements, the generation of a new view that highlights the discrepancies between the two models corresponding to the mediator pattern should conform to: R1 - highlighting the relationship $A \xrightarrow{c} B$ as being causal, and R2 - highlighting the relationships $A \rightarrow M$ and $M \rightarrow B$ as also being causal. We formally elaborate the realization of step \#3 in our method (sec \ref{sec:method}) for the mediator pattern in algorithm~\ref{alg:sequential-structure-3}.}
\chadded[id=addr2]{Algorithm~\ref{alg:sequential-structure-3} is relevant when the LiNGAM output DAG reveals a mediator structure (with $M$ as the mediator), whereas the PD algorithm identifies this structure as the sequence $A \rightarrow M \rightarrow B$. To reconcile this discrepancy for the mediator pattern, the generated view should adhere to: R1 - marking the relationship $A \xrightarrow{c} B$ as causal, and R2 - confirming that both $A \rightarrow M$ and $M \rightarrow B$ are causal. Step \#3 for the mediator pattern is elaborated in algorithm~\ref{alg:sequential-structure-3} (see sec \ref{sec:method}).}

\begin{algorithm}
\caption{Given a \chdeleted[id=del]{PDM}\chadded[id=add]{discovered BP} and a corresponding CBP that may consist of a MEDIATOR pattern}
\label{alg:sequential-structure-3}
\begin{algorithmic}[1] 
\Require 
    \State $G_{BP},G_L$ \Comment{Given BP and CBP graphs}
    \State $\forall t \in G_{BP} \implies \exists t' \in G_L$ \Comment{$t$ and $t'$ denote the same task in the respective models}
    \State $G_{result}\gets G_{BP}$ \Comment{$G_{result}$ highlights the $\Delta(G_{BP}-G_L)$}
    \ForAll {$(a,m,b) \in G_{BP}:a \rightarrow m, m \rightarrow b, a \not\rightarrow b$}
        \If{$\exists(a',m',b') \in G_L: a' \xrightarrow{c} m', m' \xrightarrow{c} b', a' \xrightarrow{c} b'$} \Comment{Check $m'$ is a mediator in $G_L$:}
            \State $G_{result}\gets$ \Call{$G_{BP}$.label}{$a \rightarrow m$, `C'}
            \State $G_{result}\gets$ \Call{$G_{BP}$.label}{$m \rightarrow b$, `C'}
            \State $G_{result}\gets$ \Call{$G_{BP}$.add}{$a \xrightarrow{c} b$}
        \EndIf
    \EndFor
    \State \textbf{return} $G_{result}$
\end{algorithmic}
\end{algorithm}

\chdeleted[id=del]{Algorithm~\ref{alg:sequential-structure-3} applies to the case where the LiNGAM output DAG shows a mediator structure (i.e., being $C$ the collider) while the PD algorithm discovers this structure as a sequence of $A \rightarrow M \rightarrow C$.} \chdeleted[id=delr2]{We note that we also check for $A \not\rightarrow B$ since such a sequence gives rise to the possibility of $M$ not being executed at all in the discovered BP. Provided $M$ is a mediator in this case, a variant in which $M$ doesn't execute is not feasible.
In this case, the algorithm sets the resulting graph $G_{result}$ to initially match the sequential structure in the PD, adds the missing causal edge $A \xrightarrow{c} B$, and explicitly annotates the directly-follows edges $A \rightarrow M$ and $M \rightarrow B$ with a `$c$' symbol to indicate their essence of being causal.}
\chadded[id=addr2]{For cases involving a mediator, we also verify the absence of $A \not\rightarrow B$, as this sequence may prevent $M$ from executing in the discovered BP. When $M$ functions as a mediator, any variant omitting $M$ is not feasible. Here, the algorithm sets $G_{result}$ to match the PD's sequential layout, adds the missing causal edge $A \xrightarrow{c} B$, and marks the directly-follows edges $A \rightarrow M$ and $M \rightarrow B$ with `$c$' symbols to denote their causal roles.}

\subsection*{\textbf{Step IV: Construct a new view that combines the results in step 3 to highlight all the discrepancies}}

\chdeleted[id=del]{Each of the three algorithms applies in a situation where the PDM deviates from the CBP structure. However, to}\chadded[id=add]{To} accommodate for any situation where the \chdeleted[id=del]{PDM}\chadded[id=add]{BP model} is consistent with the CBP structure, we also need to annotate all remaining \texttt{directly-follows} edges in the \chdeleted[id=del]{PDM}\chadded[id=add]{BP model} that are also determined as causal execution dependencies in the CBP. Thus, we annotate as causal any remaining edge $(a,b) \in G_{BP}: a \rightarrow b$ for which $(a',b') \in G_L: a' \xrightarrow{c} b'$. Lastly, we unify all the results of all three algorithms jointly with the annotations into a single $G_{result}$ graph.

\chdeleted[id=del]{We applied our method first using a generated process log with timestamps that adhere to the three execution patterns, and secondly using two real benchmark datasets~\cite{Sepsisdata2016}~\cite{BPI2020}.}

%% file: cpm4sax-exp-results.tex
\section{\texorpdfstring{\chdeleted[id=del]{Experimental setup and evaluation}\chadded[id=add]{Soundness and applicability evaluation of our method}}{Soundness and applicability evaluation of our method}}
\label{sec:evaluation}




In order to corroborate our hypothesis, i.e., the ability to form a faithful CBP \chadded[id=add]{model} and highlight potential discrepancies w.r.t the \chdeleted[id=del]{counterpart} \chdeleted[id=del]{PDM}\chadded[id=add]{BP model} \chadded[id=add]{counterpart}, 
we carried out as a first step several experiments related to our loan example in Fig~\ref{fig:bp}, using simulated data and two PD algorithms: \emph{alpha miner} and \emph{heuristic miner}, as detailed in the following subsections. The synthetic datasets can be accessed here\footnote{\url{https://github.com/IBM/SAX/tree/main/CPD2023}}. 

\chadded[id=add]{The soundness of our method was assessed with respect to the two aforementioned requirements R1 and R2.}
\chadded[id=add]{For the sake of applicability evaluation}\chdeleted[id=del]{Subsequently}, we also complemented the synthetic evaluation with two \chdeleted[id=del]{realistic}\chadded[id=add]{real-world} benchmark datasets: Sepsis~\cite{Sepsisdata2016} and Request-for-payment~\cite{BPI2020}. \chadded[id=add]{As proof of applicability, two concrete types of method applications are presented: discrepancy analysis between the discovered BP and the CBP, and comparative analysis of CBPs among different process variants. In this work, focused on establishing the feasibility of our method, we acknowledge that it remains susceptible to other qualitative evaluation aspects, including usability and performance, which will be empirically assessed in the future.}

\chadded[id=add]{With regard to the theoretical performance of our method, two specific implementations of LiNGAM are available~\protect\footnote{\url{https://sites.google.com/view/sshimizu06/lingam}}: DirectLiNGAM and ICALiNGAM. Similar to existing discovery techniques, the performance of both is directly related to the number of variables (i.e., tasks) whose interrelationships are to be examined. Specifically, DirectLiNGAM uses a direct regression approach with a computational complexity that is linear in the number of tasks.} \chadded[id=addr2]{For example, for a variant with 35 traces and 3 tasks in the Sepsis benchmark dataset, the runtime was 0.7 seconds, and for a variant with 350 traces and 5 tasks in the Request-for-Payment dataset, the runtime was 1.8 seconds.} 

\chadded[id=add]{However, DirectLiNGAM strictly requires non-Gaussian distributions. ICALiNGAM is less strict in this regard, but it is more computationally intensive. It employs an independent component analysis (ICA) based optimization step, which is polynomial (quadratic or cubic) in the number of tasks, and a matrix permutation step with factorial complexity, giving it an overall exponential complexity. Thus, as long as the non-Gaussian distribution assumption holds, DirectLiNGAM is preferable. Alternatively, when using ICALiNGAM is necessary, the computation time ranges from seconds to minutes for 10 to 100 tasks but increases exponentially as the number of tasks grows. Typically, business processes are less likely to contain hundreds of activities, so its practical applicability is still feasible. Beyond the LiNGAM implementation, the complexity of our method is also quadratic in the number of tasks due to the pairwise comparisons in step II. Similarly, the Alpha and Heuristic algorithms generally exhibit quadratic complexity ($O(n^2)$) in the number of process activities, determined by analyzing directly-follows relationships.}

To test our hypothesis, we ran each of the datasets with \chadded[id=add]{Direct}LiNGAM. 
In all cases, timestamps were adjusted according to the \texttt{absolute} modality with respect to \chadded[id=add]{the} process start time. \chadded[id=add]{For comparison between the two modalities, we also employed the \texttt{relative} modality to the Request-for-payment data.} 

\subsection{\chadded[id=add]{Soundness and applicability e}\chdeleted[id=del]{E}valuation with synthetic data}

In the following subsections, we describe our evaluation for each of the patterns using a series of generated event logs corresponding to different fragments of the BP model depicted in Figure~\ref{fig:bp}. In all three patterns, data generation adhered to a set of pattern-specific equations, reflective of the assumptions of linear relationships and error independence.  Each log was generated consisting of 9999 cases using the BIMP open-source log simulation tool \footnote{\url{https://bimp.cs.ut.ee/simulator/}}\chadded[id=addr2]{, which is the maximum number of cases the BIMP simulator can generate}. The generated data included activity times drawn from either uniform or exponential distributions, with start times every 5 and 30 minutes respectively. Specific configurations for each log distribution are elaborated further in each pattern. 

\subsubsection{Confounder}

We conducted a total of 10 experiments for the confounder pattern, having five logs in each experiment. 
Data generations adhered to the case where the $Accept$ task is a confounder, following the equations:

\setlength{\abovedisplayskip}{3pt}
\setlength{\belowdisplayskip}{3pt}
\begin{equation}
\begin{aligned}
\label{eqn:simulated-tasks-confounder}
    & Accept_{time} = Start_{time} + D_1 \\ 
    & Email_{time} = Accept_{time} + D_2 \\
    & Archive_{time} = Accept_{time} + D_3
\end{aligned}
\end{equation}

We have created two datasets conforming to the configuration shown in Table~\ref{tab:configuration-confounder}. 
In the first \chadded[id=add]{one}, $Email$ always completes before $Archive$ has started, dictating full-time precedence between the two activities. In the second \chadded[id=add]{one}, we introduced \chadded[id=add]{an} exponential distribution in both activities, assuring \chadded[id=add]{that} $Email$ completes before $Archive$ in most of the cases. 
Intervals between start times \chadded[id=add]{(relative to $Accept_{time}$ as in Equation~\ref{eqn:simulated-tasks-confounder})} were extended in the exponential case 
to ensure a small amount of swapping \chdeleted[id=del]{cases}\chadded[id=add]{activities between $Email$ and $Archive$}. \chadded[id=add]{Swapping means that the execution order of the two activities alternates between different traces.}
Data generation was conforming to the assumptions of a known confounder \chadded[id=add]{as stated in Section~\ref{sec:causal-patterns}}. 

\begin{table}[ht]
\centering
\caption{Configuration of random duration variables (in sec.) for each event type.}
\scalebox{1}
{
\label{tab:configuration-confounder}
\begin{tabular}{|c|c|c|}
\hline
\textbf{Event} & \textbf{Uniform} & \textbf{Exponential} \\
\textbf{name} & \textbf{duration} & \textbf{duration} \\
\hline
Accept  & $D_1\sim\mathcal{U}$([2,4]) & $D_1\sim$Exp(2) \\
\hline
Email  & $D_2\sim\mathcal{U}$([7,9]) & $D_2\sim$Exp(2)\\
\hline
Archive & $D_3\sim\mathcal{U}$([10,12]) & $D_3\sim$Exp(60)\\
\hline
\end{tabular}
}
\end{table}

\subsubsection{Collider}

We conducted two experiments for the collider pattern. 
Specifically, data generations adhered to the case where the $CloseApplication$ task is a collider, following the equations:

\setlength{\abovedisplayskip}{3pt}
\setlength{\belowdisplayskip}{3pt}
\begin{equation}
\begin{aligned}
\label{eqn:simulated-tasks-collider}
    & Email_{time} = Start_{time} + D_1 \\ 
    & Archive_{time} = Start_{time} + D_2 \\
    & CloseApplication_{time} = Email_{time} + \\
    &\hphantom{{}CloseApplication_{time} = } Archive_{time} + D_3
\end{aligned}
\end{equation}

We have created two datasets conforming to the configuration shown in Table~\ref{tab:configuration-collider}. In the first \chadded[id=add]{one} \chadded[id=add]{(column \#2)}, $Email$ always completes before $Archive$ has started \chadded[id=add]{as drawn from non-overlapping distributions}, dictating full-time precedence between the two activities. In the second \chadded[id=add]{one} \chadded[id=add]{(column \#3)}, also \chadded[id=add]{having} a uniform distribution in both activities \chadded[id=add]{with overlapping distributions}, assuring $Email$ completes before $Archive$ in \chadded[id=add]{about half}\chdeleted[id=del]{most} of the cases.

\begin{table}[ht]
\centering
\caption{Configuration of random duration variables (in sec.) for each event type.}
\scalebox{1}
{
\label{tab:configuration-collider}
\begin{tabular}{|c|c|c|}
\hline
\textbf{Event} & \textbf{Uniform} & \textbf{Uniform} \\
\textbf{name} & \textbf{duration} & \textbf{duration} \\
\hline
Email  & $D_1\sim\mathcal{U}$([5,7]) & $D_1\sim\mathcal{U}$([5,9])\\
\hline
Archive & $D_2\sim\mathcal{U}$([9,11]) & $D_2\sim\mathcal{U}$([7,11])\\
\hline
CloseApplication  & $D_3\sim\mathcal{U}$([2,4]) & $D_3\sim\mathcal{U}$([2,4]) \\
\hline
\end{tabular}
}
\end{table}

\subsubsection{Mediator}

We conducted an additional experiment for the mediator pattern. 
Data generations adhered to the case where the $CloseApplication$ task is a partial mediator for the effect the $Archive$ activity has on the $PaperDisposal$ activity, following the equations:

\setlength{\abovedisplayskip}{3pt}
\setlength{\belowdisplayskip}{3pt}
\begin{equation}
\label{eqn:simulated-tasks-mediator}
\begin{aligned}
    & Archive_{time} = Start_{time} + D_1  \\ 
    & CloseApplication_{time} = Archive_{time} + D_2  \\
    & PaperDisposal_{time} = CloseApplication_{time} + \\  
    &\hphantom{{}PaperDisposal_{time} = } Archive_{time} + D_3
\end{aligned}
\end{equation}

In the created dataset, all activity durations were drawn from a uniform distribution as shown in Table~\ref{tab:configuration-mediator}.

\begin{table}[ht]
\centering
\caption{Configuration of random duration variables (in sec.) for each event type.}
\scalebox{1}
{
\label{tab:configuration-mediator}
\begin{tabular}{|c|c|}
\hline
\textbf{Event} & \textbf{Uniform}  \\
\textbf{name} & \textbf{duration}  \\
\hline
Archive  & $D_1\sim\mathcal{U}$([7,11]) \\
\hline
CloseApplication & $D_2\sim\mathcal{U}$([5,7]) \\
\hline
PaperDisposal  & $D_3\sim\mathcal{U}$([2,4]) \\
\hline
\end{tabular}
}
\end{table}

\subsection{\chadded[id=add]{Soundness and applicability e}\chdeleted[id=del]{E}valuation with benchmark data}

We explored two benchmark datasets: Sepsis and Request-for-payment. 
The former shows an example of the difference between the \chdeleted[id=del]{PM}\chadded[id=add]{discovered BP} and CBP \chadded[id=add]{models} as \chdeleted[id=del]{may} arise\chadded[id=add]{s from} \chdeleted[id=del]{in realistic}\chadded[id=add]{real-world} datasets, and the latter shows the benefit of \chadded[id=add]{comparing}\chdeleted[id=del]{using complementary} causal perspectives associated with different process variants when trying to explain certain outcomes.

\subsubsection{Discrepancy between \chdeleted[id=del]{PDM}\chadded[id=add]{BP} and CBP \chadded[id=add]{models}}

To complement the synthetic evaluation, we investigated the hospital Sepsis benchmark dataset. This open dataset \chadded[id=add]{holds records of 1050 patients experiencing sepsis symptoms}\chdeleted[id=del]{caters to an event log for a sample of 1050 patients with symptoms of a sepsis condition, recording their experience}, as captured by the ERP system of the hospital, from \chadded[id=add]{their} arrival in the emergency ward to \chadded[id=add]{their} admission to the hospital. Overall, the event log consists of a total of 15,000 events that were recorded for 16 different activities. Our particular focus using this dataset has been on identifying a confounder-related situation that presents inconsistency compared to the corresponding process mining structure for further investigation of the discrepancy and possible insight derivation. 

Given the inherent complexity and variability in health care processes, in the case of the Sepsis dataset, we employed the heuristic miner algorithm, and the IBM Process Mining tool\footnote{\url{https://www.ibm.com/products/process-mining}}. We did not use the $\alpha$ algorithm due to its limitation when applying it to noisy or complex data \cite{Parente2022}.
As with the synthetic data, LiNGAM was then employed to elicit a corresponding CBP perspective. 

\subsubsection{Comparative analysis between variants}

We explored a part of the benchmark BPI2020 to elaborate on the value of the causal execution view for comparative analysis of different variants. 
Overall, this dataset contains events related to two years of travel expense claims for domestic and international trips by university employees from 2017 to 2018.
Concretely, we inspected the part of the dataset that contains events related to Requests for Payment that include 6,886 cases, and 36,796 events. We chose to investigate the condition where the payment request is rejected by the administrator which contains 755 cases overall. Following such a rejection, the process unfolds in one of three prominent variants.

For the sake of comparative analysis between variants, we compared the IBM Process Mining tool and LiNGAM results.

\section{Results}
\label{sec:results}

\subsection{Synthetic data results}

Our findings are consistent with our hypothesis. Across all three patterns, for the cases with no swapping, the applied PD algorithms results \chdeleted[id=del]{(i.e., the PDMs)} did \chdeleted[id=del]{n't}\chadded[id=add]{not} match the CBP structure. \chdeleted[id=del]{This also applies when the swapped cases have been removed from the event log.}


\subsubsection{Confounder}

\begin{figure*}[htbp]
\centering
\begin{tabular}{cc}
\parbox[c]{0.5\hsize}{\centering\includegraphics[scale = 0.22]{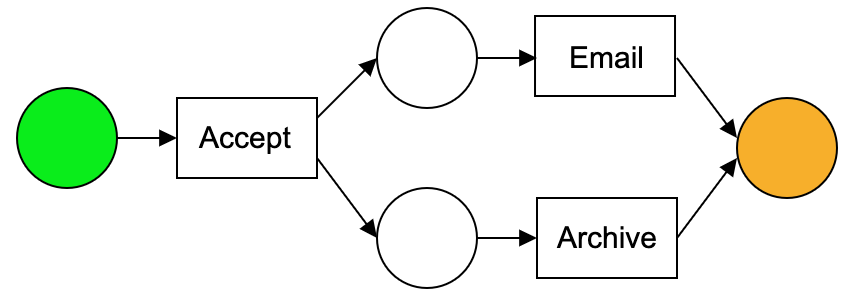}}
& 
\parbox[c]{0.5\hsize}{\centering\includegraphics[scale = 0.22]{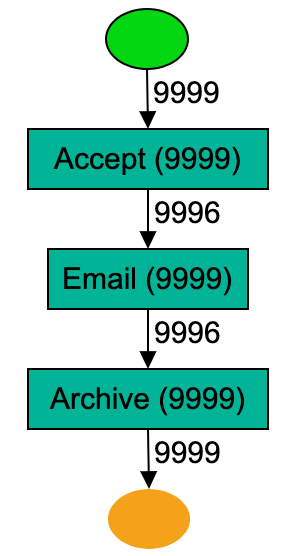}}     
\\
(a) $\alpha$-algorithm & (b) Heuristic algorithm
\\
\noalign{\vskip 1mm} 
\end{tabular}%
\caption{Process discovery results for the exp. case with no removal of swapped cases in the confounder pattern.}
\label{fig:exp-case-pm-results}
\end{figure*}

We first show the results corresponding to the synthetic datasets configurations in Table~\ref{tab:configuration-confounder}, split by distribution. For the uniform case, in both algorithms, results were consistent across all five logs. The respective process mining algorithm results for each of the log runs are illustrated in Fig~\ref{fig:uniform-case-pm-results}. For the exponential case, the results are shown in Fig~\ref{fig:exp-case-pm-results}.

Note that in the exponential case, in each log there were several execution traces in which the order of the $Email$ and the $Archive$ tasks is reversed (\chadded[id=add]{this number}{} ranging \chadded[id=add]{between} 3 \chadded[id=add]{to}\chdeleted[id=del]{-} 15). As a result, the $\alpha$-algorithm always concluded a split after the execution of the $Accept$ task, subsequently with no deterministic order between the $Email$ and $Archive$ tasks. Structure-wise, this result is, in fact, equivalent to the CBP model. On the contrary, being more robust to order swapping, the Heuristic algorithm yielded the same sequential flow structure, where the numbers on the edges denote the number of instances that conform to the execution order (i.e., excluding swapped occurrences). Specifically, Fig~\ref{fig:exp-case-pm-results}(b) shows the result for a log where the number of swapped cases equals \chadded[id=add]{to} three.

\begin{figure*}[htbp]
\centering
\begin{tabular}{cc}
\parbox[c]{0.5\hsize}{\centering\includegraphics[scale = 0.22]{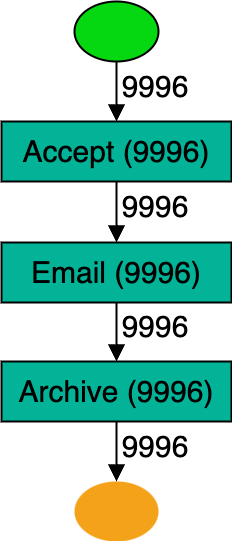}}
& 
\parbox[c]{0.5\hsize}{\centering\includegraphics[scale = 0.22]{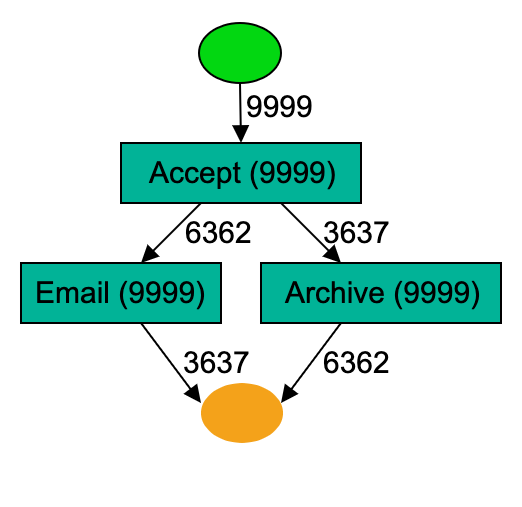}}     
\\
(a) Heuristic algorithm with & (b) Heuristic algorithm with\\ three swapped cases removed.  & significant swapping cases.
\\
\noalign{\vskip 1mm} 
\end{tabular}%
\caption{Sensitivity of the Heuristic algorithm to swapped cases in the confounder pattern.}
\label{fig:heuristic-sensitivity}
\end{figure*}

To test for the sensitivity of the Heuristic algorithm to the number of swapped cases, we ran two additional tests, one in which we filtered out the swapped cases, and a second in which we generated a log with the uniform distribution configuration, but altered the duration for the $Archive$ task to be drawn from $\mathcal{U}[8,10]$, to ensure it significantly overlaps with the $Email$ task. The corresponding results for these two tests are presented in Fig~\ref{fig:heuristic-sensitivity}. Running a similar test with the $\alpha$-algorithm was skipped, given its rigid sensitivity to the number of swapped cases that consistently leads to the result in Fig.~\ref{fig:exp-case-pm-results}(a).

\begin{figure*}[htbp]
\centering
\begin{tabular}{ccc}
\parbox[c]{0.30\hsize}{\centering\includegraphics[scale = 0.25,clip,trim={0 0.9cm 0 1.3cm}]{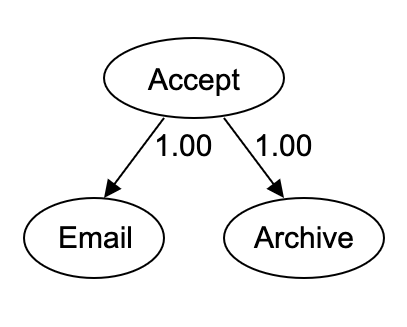}}
& 
\parbox[c]{0.30\hsize}{\centering\includegraphics[scale = 0.25,clip,trim={0 0.9cm 0 1.3cm}]{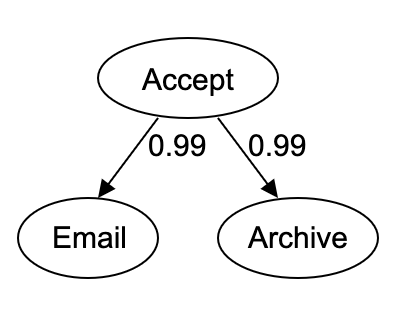}} 
&
\parbox[c]{0.30\hsize}{\centering\includegraphics[scale = 0.25,clip,trim={0 0.9cm 0 1.3cm}]{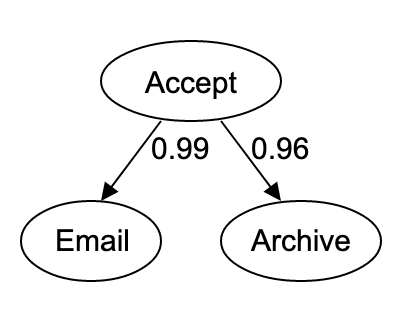}} 
\\
(a) LiNGAM algorithm with & (b) LiNGAM algorithm with & (c) LiNGAM algorithm with \\ no swapping cases.  & few swapping cases. & many swapping cases.
\\
\noalign{\vskip 1mm} 
\end{tabular}%
\caption{LiNGAM algorithm results for the confounder pattern.}
\label{fig:lingam-results-confounder}
\end{figure*}

As with the case of having only a few swapped cases, their elimination retained a sequential structure, as illustrated in Fig~\ref{fig:heuristic-sensitivity}(a).
However, for the second test, we observed that the significant number of swaps also affected the result of the Heuristic algorithm, yielding a structure illustrated in Fig~\ref{fig:heuristic-sensitivity}(b) that corresponds to the CBP as illustrated in Fig~\ref{fig:lingam-results-confounder}. 

Given the above results, we tested the LiNGAM algorithm and
assessed its results in comparison to each of the cases generated by the \chdeleted[id=del]{PM}\chadded[id=add]{PD} algorithms, as summarized in Fig~\ref{fig:lingam-results-confounder}.

\begin{itemize}[topsep=2pt,leftmargin=*]
\item Fig~\ref{fig:lingam-results-confounder}(a) relates to the cases where there were no swaps at all (\chadded[id=add]{as in} Fig~\ref{fig:uniform-case-pm-results}(a) and~\ref{fig:uniform-case-pm-results}(b)), and also to the case where the few swaps were removed from the event logs as in Fig~\ref{fig:heuristic-sensitivity}(a).
\item Fig~\ref{fig:lingam-results-confounder}(b) relates to the cases where there was a limited number of swapped cases with no removal, as in Fig~\ref{fig:exp-case-pm-results}(a) and \ref{fig:exp-case-pm-results}(b).
\item Fig~\ref{fig:lingam-results-confounder}(c) relates to the case with significant swapping cases as in Fig~\ref{fig:heuristic-sensitivity}(b).
\end{itemize}

For the cases with a few swapped instances, the $\alpha$ algorithm generated a flow structure that matches the CBP, whereas the Heuristic algorithm generated a sequential flow structure. This stems from the difference in the threshold used by the two algorithms, with $\alpha$ having no tolerance for the presence of swapped cases. Once the amount of swapped cases exceeded the threshold, the Heuristic algorithm also generated a flow structure that matched the confounder pattern. 
In all these cases, LiNGAM was consistently able to identify the confounder pattern.  Particularly, we can conclude that LiNGAM was able to correctly recognize the cases in which the \chdeleted[id=del]{PM}\chadded[id=add]{PD} algorithms generated the sequential pattern $A \rightarrow B  \rightarrow C$ and that regression coefficients, as denoted on the edges of LiNGAM outputs, reflect the tendency towards the true value of 1 as in the original data generation equations (see Equation \ref{eqn:simulated-tasks-confounder}). 

\begin{figure*}[htbp]
\centering
\includegraphics[width = 0.5\hsize]{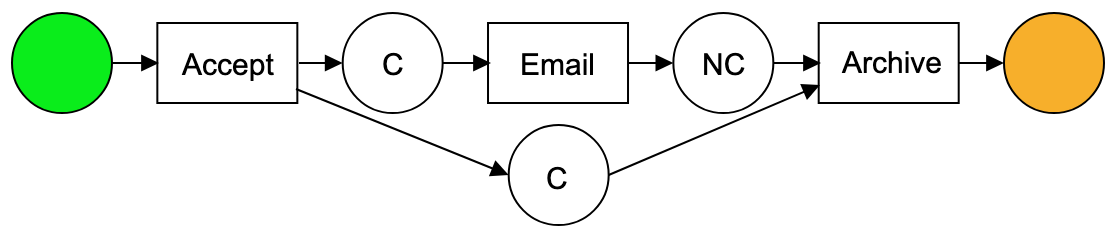}
\caption{Highlighted \chdeleted[id=del]{result} view \chadded[id=add]{($G_{result}$)} for the confounder pattern. \chadded[id=add]{Respective to the discovered BP in Fig~\ref{fig:uniform-case-pm-results}(a), the $Accept \rightarrow Email$ relation is annotated as causal (denoted by `C' inside the connecting place), the $Email \rightarrow Archive$ relation is annotated as non-causal, and the $Accept \rightarrow Archive$ relation is new and annotated as causal.}}
\label{fig:petri-corrected}
\end{figure*}

In cases when the CBP \chadded[id=add]{model} does \chdeleted[id=del]{n't}\chadded[id=add]{not} match the \chdeleted[id=del]{PDM}\chadded[id=add]{BP model}, \chadded[id=add]{the} application of algorithm~\ref{alg:sequential-structure-1} to the \chdeleted[id=del]{PMD}\chadded[id=add]{discovered BP} as in Fig~\ref{fig:uniform-case-pm-results}(a) yields the highlighted result view ($G_{result}$) as shown in Fig~\ref{fig:petri-corrected}.

\subsubsection{Collider}

\begin{figure*}[htbp]
\centering
\begin{tabular}{cc}
\parbox[c]{0.6\hsize}{\centering\includegraphics[scale = 0.22]{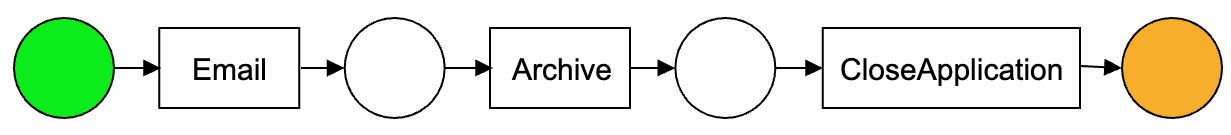}}
& 
\parbox[c]{0.4\hsize}{\centering\includegraphics[scale = 0.22]{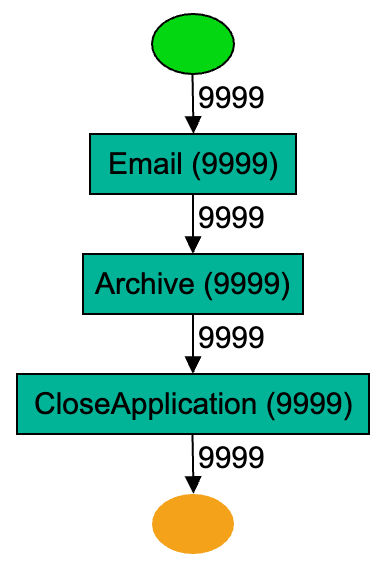}}     
\\
(a) $\alpha$-algorithm & (b) Heuristic algorithm
\\
\noalign{\vskip 1mm} 
\end{tabular}%
\caption{Process discovery results for the uniform distribution with no overlapping cases in the collider pattern.}
\label{fig:uni1-collider-case-pm-results}
\end{figure*}

\begin{figure*}[htbp]
\centering
\begin{tabular}{cc}
\parbox[c]{0.6\hsize}{\centering\includegraphics[scale = 0.22]{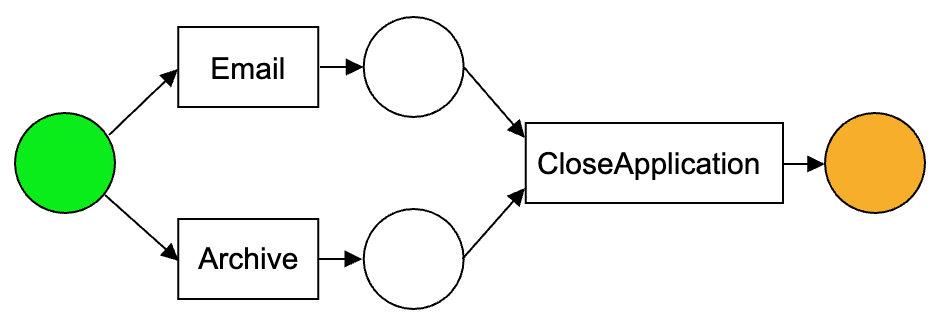}}
& 
\parbox[c]{0.4\hsize}{\centering\includegraphics[scale = 0.22]{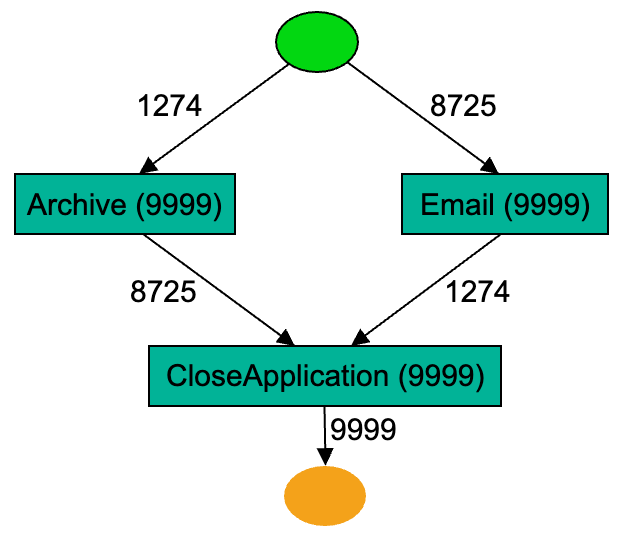}}     
\\
(a) $\alpha$-algorithm & (b) Heuristic algorithm
\\
\noalign{\vskip 1mm} 
\end{tabular}%
\caption{Process discovery results for the uniform distribution with some overlapping cases in the collider pattern.}
\label{fig:uni2-collider-case-pm-results}
\end{figure*}

Complementing the robustness tests employed for the confounder pattern, we pursued the assessment for the collider pattern with two datasets with activity duration times drawn from a uniform distribution, one with and another without overlapping activities.
The results corresponding to the dataset configuration in Table~\ref{tab:configuration-collider} are as follows. For the first uniform distribution with no overlapping cases between $Email$ and $Archive$, the process mining results for both $\alpha$ and heuristic algorithms are as illustrated in Figure~\ref{fig:uni1-collider-case-pm-results}.
For the second uniform distribution with some overlapping cases between $Email$ and $Archive$, the process mining results for the two process mining algorithms are as illustrated in ~\ref{fig:uni2-collider-case-pm-results}.

As with the confounder case, it is apparent that when there are no swapping cases, the $\alpha$-algorithm and the heuristic one both yield a sequential structure that is inconsistent with the collider pattern structure. Having only a few swapping cases, or passing the threshold for the heuristic algorithm, yields a structure that is consistent with the collider pattern.

\begin{figure}[htbp]
\centering
\includegraphics[scale = 0.25,clip]{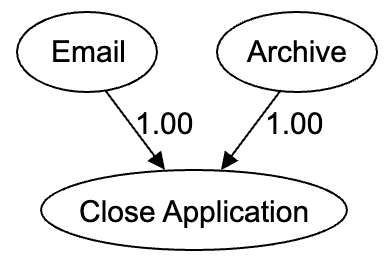}
\caption{LiNGAM algorithm results for the collider pattern.}
\label{fig:lingam-results-collider}
\end{figure}

Running the LiNGAM algorithm for the two distributions resulted in both with the same CBP structure as illustrated in Figure~\ref{fig:lingam-results-collider}. As presented, in both distributions LiNGAM was able to identify the collider pattern.

\begin{figure*}[htbp]
\centering
\includegraphics[width = 0.5\hsize]{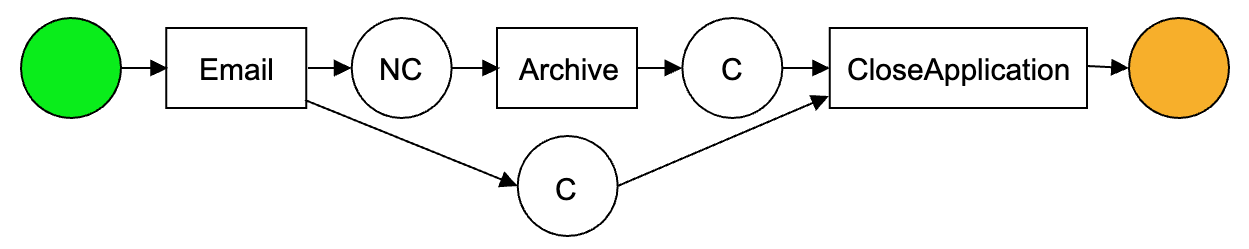}
\caption{Highlighted \chdeleted[id=del]{result} view \chadded[id=add]{($G_{result}$)} for the collider pattern. \chadded[id=add]{Respective to the discovered BP in Fig~\ref{fig:uni1-collider-case-pm-results}(a), the $Email \rightarrow Archive$ relation is annotated as non-causal (denoted by `NC' inside the connecting place), the $Archive \rightarrow CloseApplication$ relation is annotated as causal, and the $Email \rightarrow CloseApplication$ relation is new and annotated as causal.}}
\label{fig:petri-corrected-collider}
\end{figure*}

In cases when the CBP \chadded[id=add]{model} does \chdeleted[id=del]{n't}\chadded[id=add]{not} match the \chadded[id=add]{BP model}, \chadded[id=add]{the} application of algorithm~\ref{alg:sequential-structure-2} to the \chdeleted[id=del]{PMD}\chadded[id=add]{discovered BP} as in Fig~\ref{fig:uni1-collider-case-pm-results}(a) yields the highlighted result view ($G_{result}$) as shown in Fig~\ref{fig:petri-corrected-collider}.

\subsubsection{Mediator}

\begin{figure*}[htbp]
\centering
\begin{tabular}{cc}
\parbox[c]{0.6\hsize}{\centering\includegraphics[scale = 0.22]{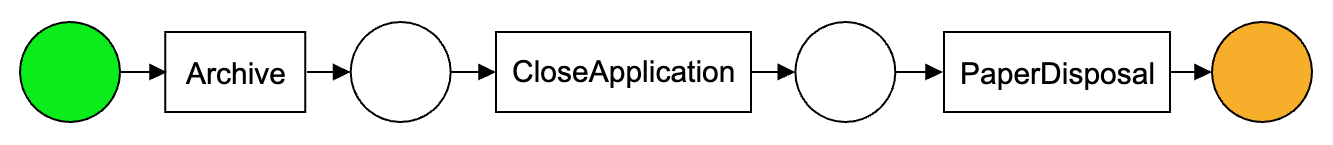}}
& 
\parbox[c]{0.4\hsize}{\centering\includegraphics[scale = 0.22]{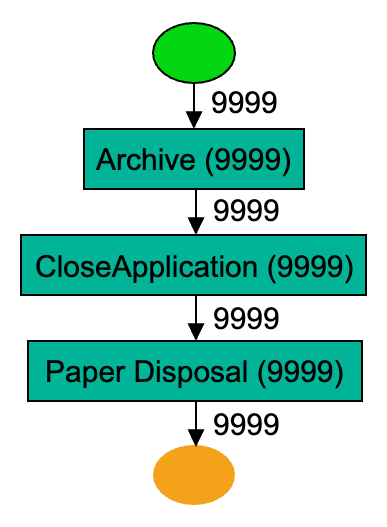}}     
\\
(a) $\alpha$-algorithm & (b) Heuristic algorithm
\\
\noalign{\vskip 1mm} 
\end{tabular}%
\caption{Process discovery results for the uniform distribution cases in the mediator pattern.}
\label{fig:uni1-mediator-case-pm-results}
\end{figure*}

We pursued the assessment for the mediator pattern with the uniform distribution dataset that corresponds to the configuration in Table~\ref{tab:configuration-mediator}. Process mining results for the $\alpha$ and heuristic mining algorithms are illustrated in Figure~\ref{fig:uni1-mediator-case-pm-results}.

As aforementioned, in the dataset that was used, the $CloseApplication$ activity acted as a partial mediator to the execution dependency between the $Archive$ and the $PaperDisposal$ activities, retaining also a direct execution dependence between the two.  
However, this direct effect was not apparent in any of the process mining results.

\begin{figure}[htbp]
\centering
\includegraphics[scale = 0.25,clip]{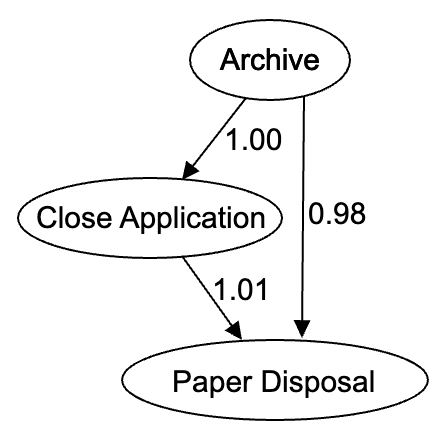}
\caption{LiNGAM algorithm results for the mediator pattern.}
\label{fig:lingam-results-mediator}
\end{figure}

Running the LiNGAM algorithm for the same dataset yielded a CBP structure as illustrated in Figure~\ref{fig:lingam-results-mediator}. As presented, LiNGAM did construct the pattern properly, capturing the mediator pattern with part of the effect being mediated and part being direct.

\begin{figure*}[htbp]
\centering
\includegraphics[width = 0.5\hsize]{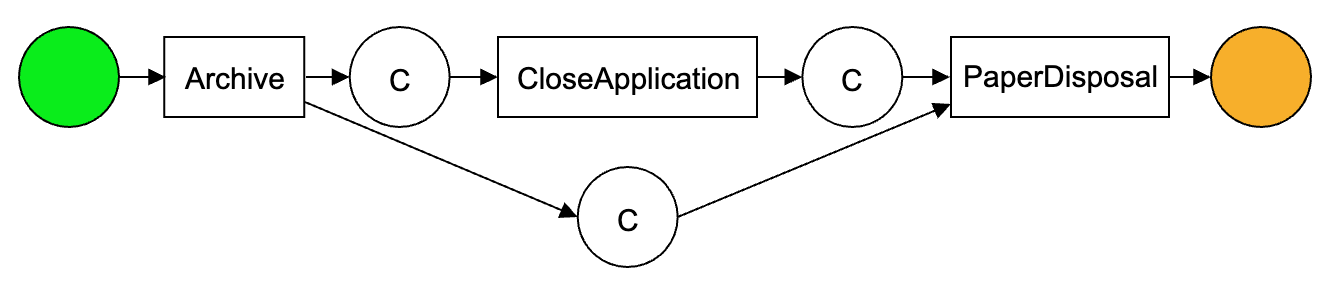}
\caption{Highlighted \chdeleted[id=del]{result} view \chadded[id=add]{($G_{result}$)} for the mediator pattern. \chadded[id=add]{Respective to the discovered BP in Fig~\ref{fig:uni1-mediator-case-pm-results}(a), the $Archive \rightarrow CloseApplication$ relation is annotated as causal (denoted by `C' inside the connecting place), the $CloseApplication \rightarrow PaperDisposal$ relation is annotated as causal, and the $Archive \rightarrow PaperDisposal$ relation is new and annotated as causal.}}
\label{fig:petri-corrected-mediator}
\end{figure*}

Application of algorithm~\ref{alg:sequential-structure-3} to the \chdeleted[id=del]{PMD}\chadded[id=add]{discovered BP} as in Fig~\ref{fig:uni1-mediator-case-pm-results}(a) yields the highlighted result view ($G_{result}$) as shown in Fig~\ref{fig:petri-corrected-mediator}.

\subsection{Benchmark data results}

\subsubsection{Discrepancy between \chdeleted[id=del]{PDM}\chadded[id=add]{BP} and CBP \chadded[id=add]{models}}

\begin{figure*}[htbp]
    \centering
    {\includegraphics[width=\textwidth]{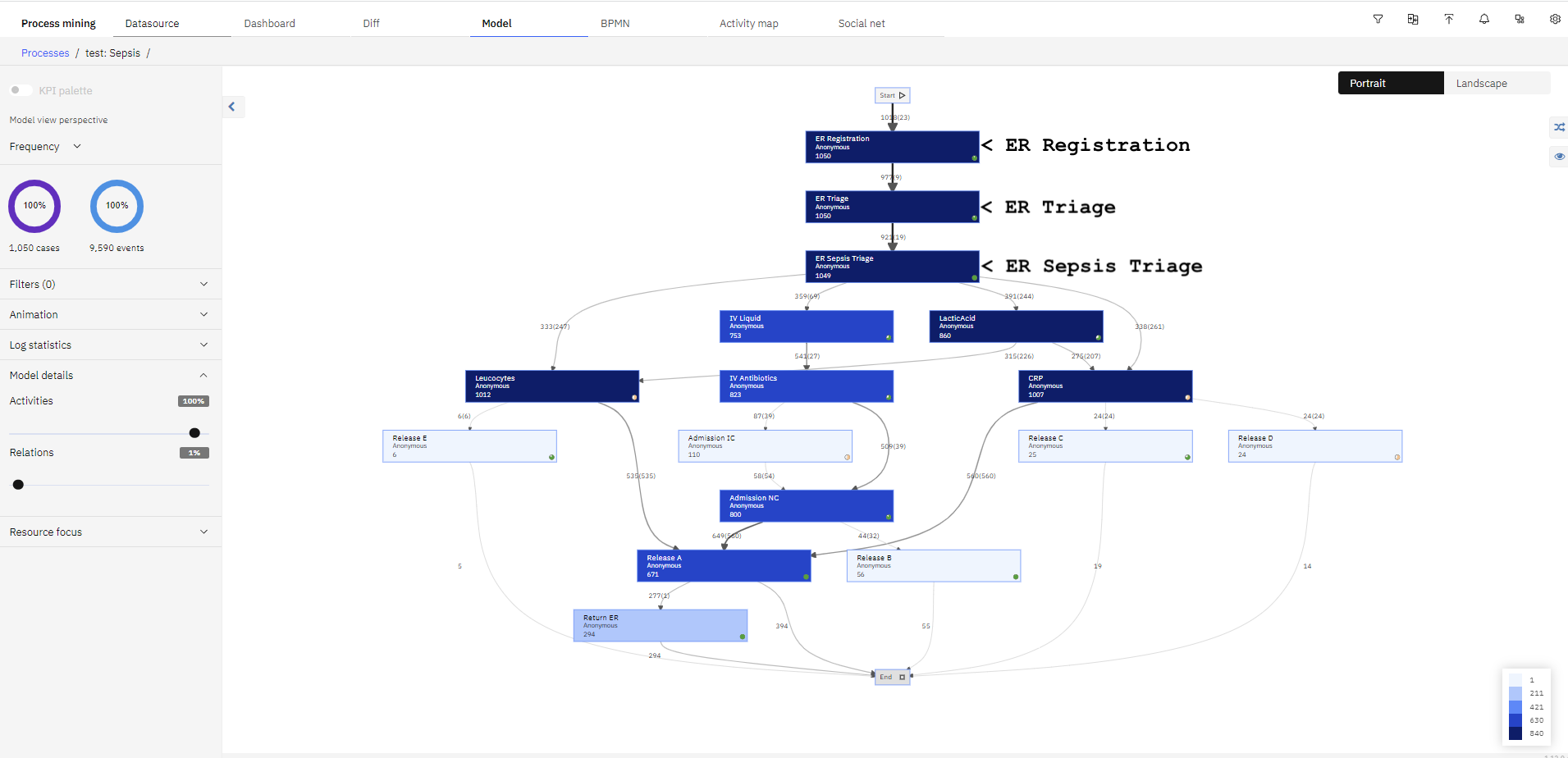}}
    \vspace*{-2mm}
    \caption{IBM Process Mining \chdeleted[id=del]{PDM}\chadded[id=add]{BP model} for the Sepsis data.}
    \label{fig:ibm-pm-sepsis}
\end{figure*}

In the second step of the evaluation, examining the hospital Sepsis dataset, we also searched for a situation where \chdeleted[id=del]{causal discovery}\chadded[id=add]{CD} uncovers a confounder execution pattern while conventional process mining yields a different structure. Both the Heuristic mining algorithm and the IBM Process Mining tool revealed a sequential process structure related to the first three activities in the process: ER Registration, ER Triage, and ER Sepsis Triage as shown in Fig \ref{fig:ibm-pm-sepsis}.

The Sepsis dataset also consists of a series of data attributes captured along the events. Among these attributes, the data about each ER Registration event also includes attributes reflective of SIRS (Systemic Inflammatory Response Syndrome) screening criteria. For Sepsis patients, screening is likely to trigger both the ER Triage and the ER Sepsis triage activities, where the completion of the former also triggers the execution of the latter. Hence, a sequential time-precedence relationship among the three activities as observed in the process mining result tells only a part of the story and requires further examination in order to determine if indeed the execution of both triage activities is in fact already determined by the SIRS screening during registration.

\chadded[id=add]{Before running LiNGAM, we checked for distribution conformance. A Shapiro-Wilks test showed that the distribution of the activities ER Sepsis Triage (W = 0.36,  p-value $<$ .01) and ER Triage (W = 0.06,  p-value $<$ .01) departed significantly from normality, thus allowing the use of DirectLiNGAM.}

\begin{figure}[htbp]
    \centering
    {\includegraphics[scale = 0.4,trim={3cm 0.8cm 3cm 0.37cm},clip]{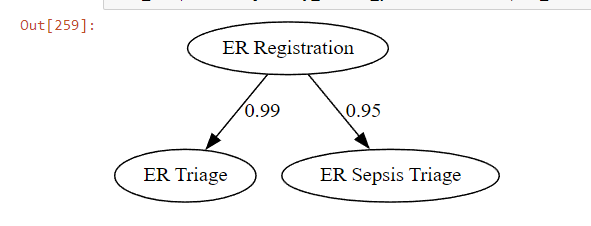}}
    \caption{CBP for the Sepsis data corresponding to the first three activities in the process.}
    \label{fig:lingam-cbp-sepsis}
\end{figure}

\begin{figure*}[htbp]
\centering
\begin{tabular}{ccc}
\parbox[c]{0.33\textwidth}{\centering\includegraphics[scale = 0.17]{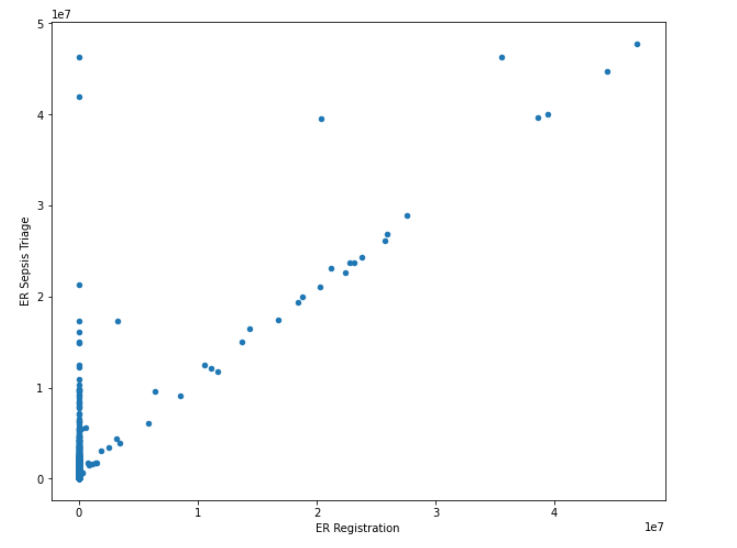}}
& 
\parbox[c]{0.33\textwidth}{\centering\includegraphics[scale = 0.17]{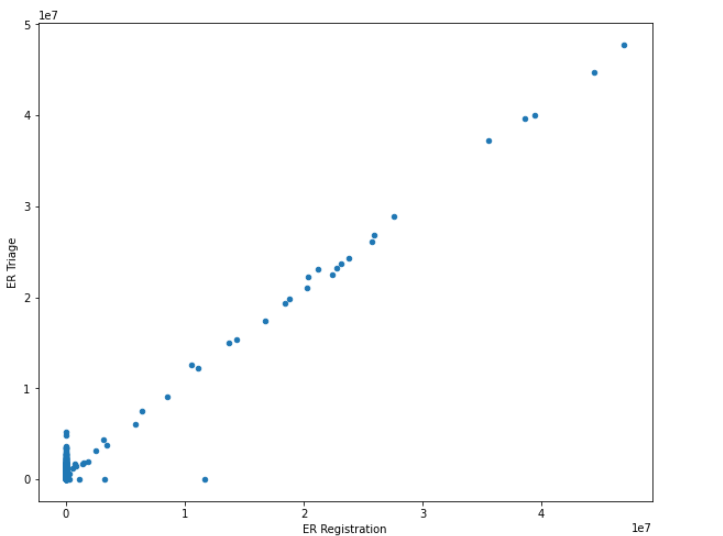}} 
&
\parbox[c]{0.33\textwidth}{\centering\includegraphics[scale = 0.17]{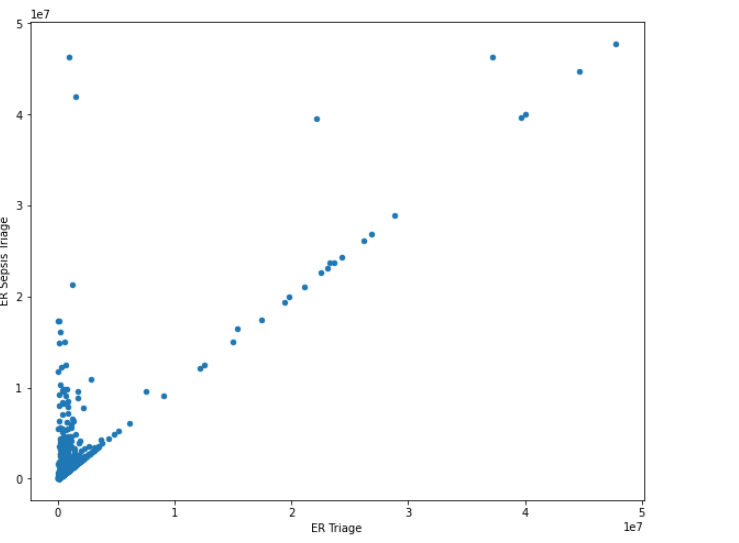}} 
\\
(a) ER Registration vs. & (b) ER Registration vs. & (c) ER Triage \\ ER Sepsis Triage  & ER Triage & ER Sepsis Triage
\\
\noalign{\vskip 1mm}  
\end{tabular}%
\caption{Correlation plots for inter-activity relationships.}
\label{fig:sepsis-edge-corr-analysis}
\end{figure*}

Elicitation of the CBP model for the same three activities yielded a confounder pattern, as shown in Fig.~\ref{fig:lingam-cbp-sepsis}. According to this result, we were able to get a confirmation that the execution of both triage activities is determined by the execution of the ER Registration activity in the process. However, the CBP did not show causal execution dependence between the basic triage and the sepsis triage. 
To investigate this lack of causal execution dependence further, we generated pair-wise XY scatter plots to observe the nature of \chadded[id=add]{the} correlations among the three activities. \chadded[id=add]{The p}\chdeleted[id=del]{P}lot results are illustrated in Fig~\ref{fig:sepsis-edge-corr-analysis}. For the two causal execution dependencies that were denoted by the arrows in the CBP, the majority of points were scattered along the diagonal of the axes in the plots, respectively, as shown in plots (a) and (b), \chadded[id=add]{in accordance to}\chdeleted[id=del]{concluding} the causal execution dependencies in Fig.~\ref{fig:lingam-cbp-sepsis}. However, the lack of causal execution dependence between the two triage activities has manifested itself in an interesting 'V'-like shape as prominently depicted in plot (c), with a significant amount of points clustered along the Y-axis. This cluster reflects process execution traces in which the execution of the sepsis triage was lagging significantly behind the execution of basic triage for cases where the latter finished relatively quickly. Such a delay is likely to be indicative of a lack of resources in the process, that is, limited availability of physicians who can attend to patients in the ER soon after the basic triage is concluded to perform the sepsis triage. In fact, but to a lesser extent, the plots also reveal a similar 'V` pattern such that for very short Registration task executions, ER Sepsis task durations vary in length. Aside from a resource issue, this could also be attributed to those instances where SIRS was not done/not done properly during registration and required repeating it during Sepsis Triage. Such insight derivation demonstrates the value of detecting and further delving into an investigation of discrepancies between the \chdeleted[id=del]{PDM}\chadded[id=add]{BP} and the CBP views.

\subsubsection{Comparative analysis between variants}

In this section, we further delved into demonstrating the value of a comparative investigation of different variants associated with a given process (intermediate) outcome. 
Using the Request-for-Payment BPI benchmark dataset, we specifically analyzed the part of the process associated with the post-rejection of a payment request by an administrator. Our analysis begins at the point where a payment request is resubmitted by the employee for approval and continues to the point where it is resubmitted and handled.

\begin{figure*}[htbp]
    \centering
    {\includegraphics[width=\textwidth]{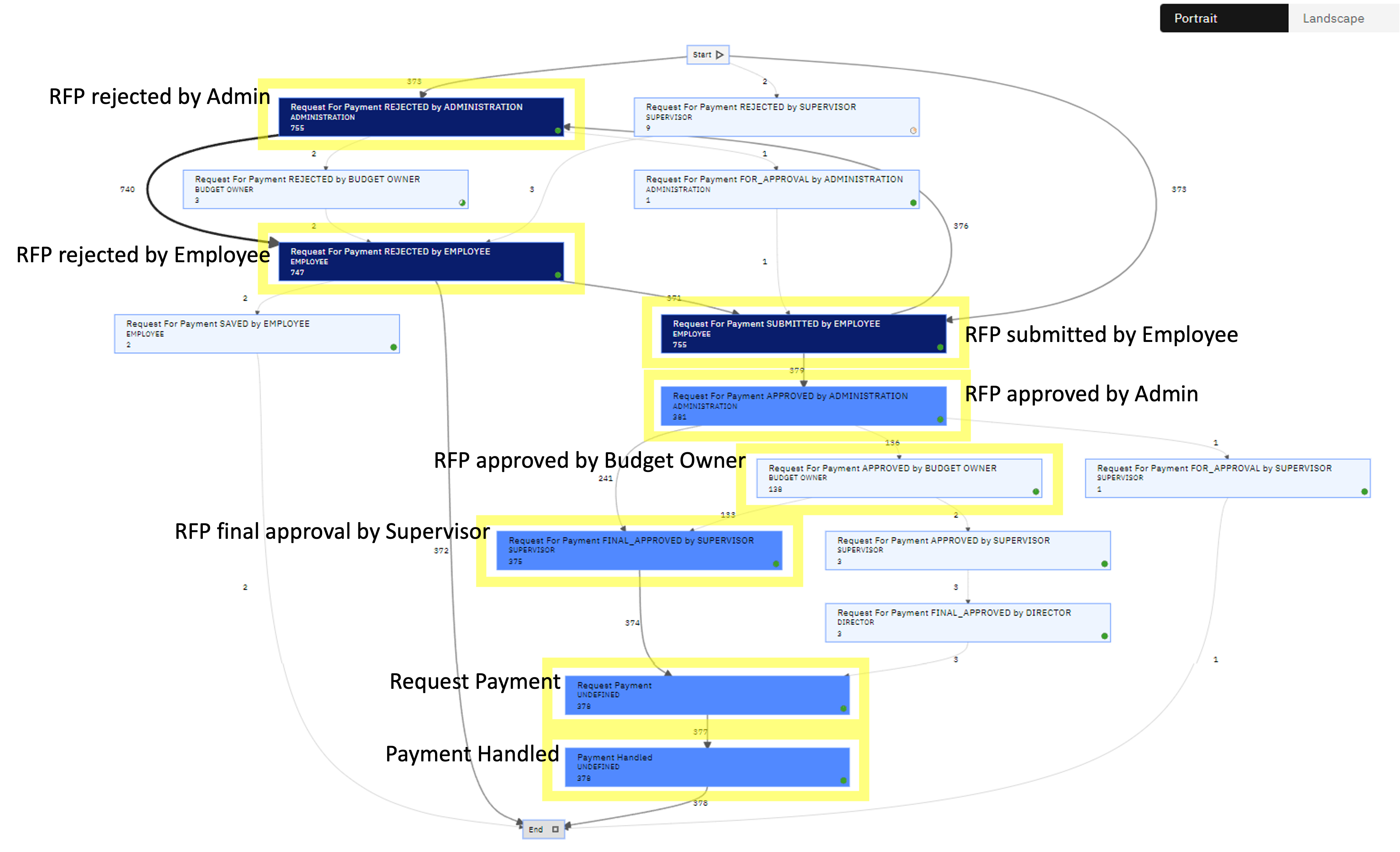}}
    \vspace*{-2mm}
    \caption{IBM Process Mining \chdeleted[id=del]{PDM}\chadded[id=add]{BP model} for the Request-for-payment dataset.}
    \label{fig:ibm-pm-payment-request}
\end{figure*}

Employing the IBM Process Mining tool for process discovery on this dataset yields the \chdeleted[id=del]{PDM}\chadded[id=add]{BP model} in Figure~\ref{fig:ibm-pm-payment-request}. We highlighted in the \chdeleted[id=del]{PDM}\chadded[id=add]{BP model} the activities that are part of the aforementioned \chdeleted[id=del]{fragment}\chadded[id=add]{segment} of interest. According to this model, following payment rejection by an Admin, it is observable that either the employee rejects the submission and withdraws from the process (reaching the process end), or \chdeleted[id=del]{that} the request is resubmitted by the employee. In this latter case, a sequence of approvals takes place, splitting between two major variants. One variant consists of 136 cases, in which the chain of approvals includes the process Admin, Budget Owner, and Supervisor. A second variant, consisting of 241 cases, skips the approval step by a Budget Owner when the Budget Owner is in fact the same person as the Supervisor. Following the approval, the payment request is repeated, and subsequently handled. In such circumstances, if there may happen to be any delays with the re-submission of the requests, the immediate activity one may choose to investigate according to the \chdeleted[id=del]{PDM}\chadded[id=add]{BP model} would be the approval-by-the-supervisor activity that precedes it.

\begin{figure*}[htbp]
    \centering
    {\includegraphics[width=\textwidth]{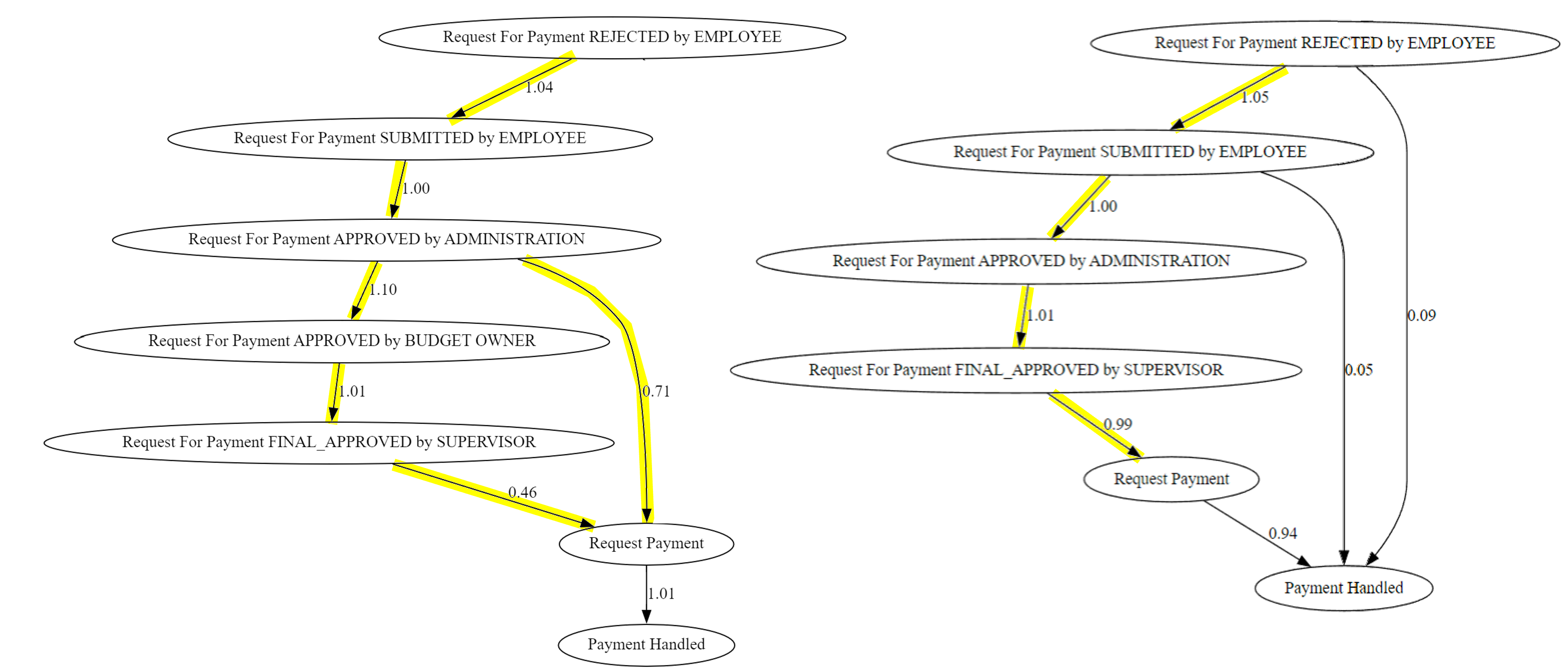}}
    \vspace*{-2mm}
    \caption{CPB comparable variants for the Request-for-payment data - \chadded[id=add]{Absolute Modality.}}
    \label{fig:CBP-payment-request}
\end{figure*}

\chadded[id=add]{The} \chdeleted[id=del]{E}\chadded[id=add]{e}licitation of the CBP for the same two major variants may  \chdeleted[id=del]{present with}\chadded[id=add]{yield} slightly different interpretations. The results of running LiNGAM for these two variants are \chdeleted[id=del]{as} illustrated in Figure~\ref{fig:CBP-payment-request}. While in the case where there are only approvals by an admin and by a supervisor, the latter activity is also the sole cause for the execution of the eventual payment request, this is not the case when there is also \chdeleted[id=del]{an} approval by a budget owner. In all such cases, the eventual $RequestPayment$ is revealed to be a collider activity, having its execution timing being also partially dependent on the initial approval by the admin. This implies that some of the delays with such requests may inherently be caused by some action that is made by the admin, and not necessarily only a responsibility of the supervisor. Such a conclusion can only be drawn from observing the CBP.

\begin{figure*}[htbp]
    \centering
    {\includegraphics[width=0.9\textwidth]{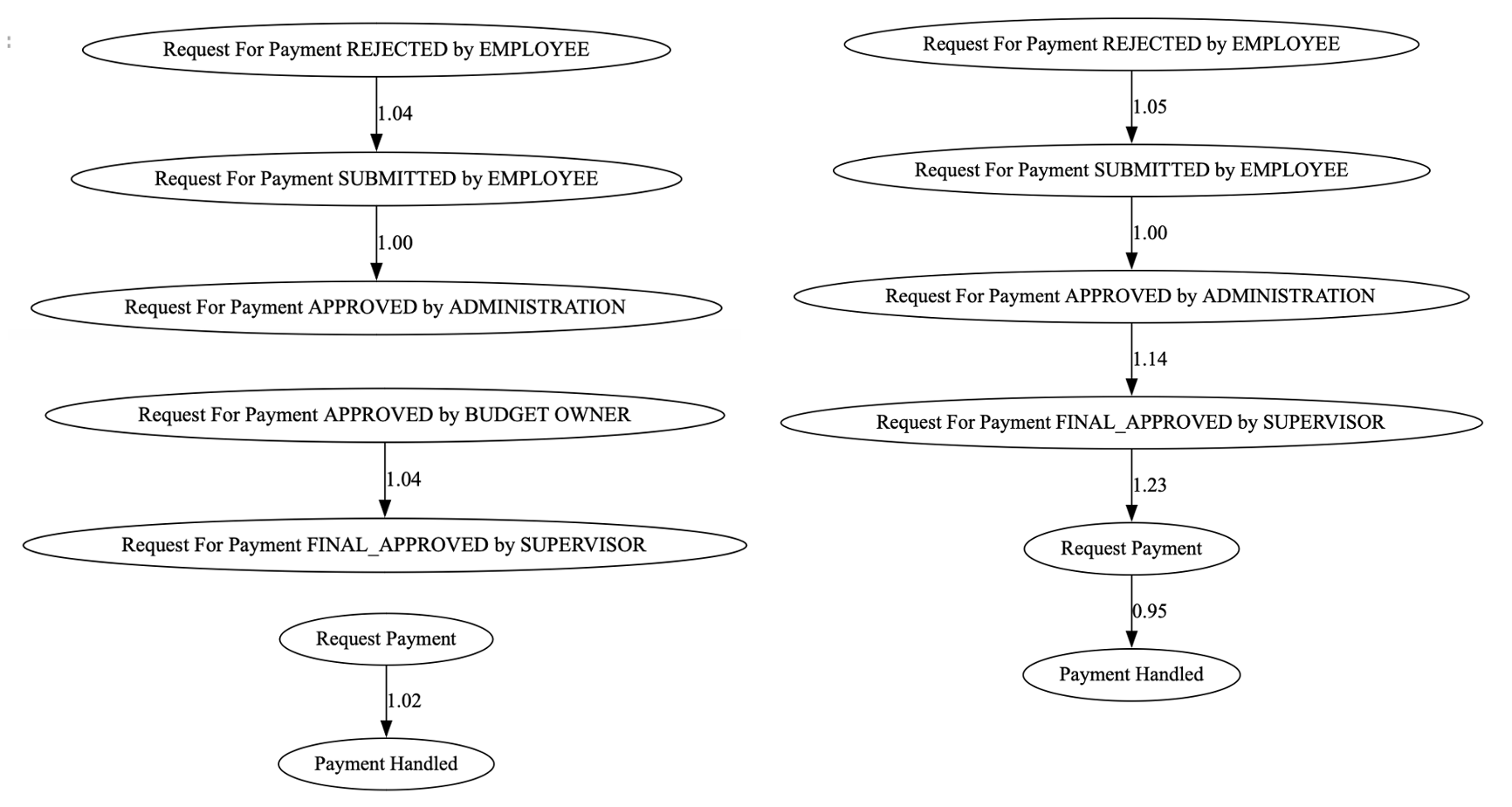}}
    \vspace*{-2mm}
    \caption{\chadded[id=add]{CPB comparable variants for the Request-for-payment data - Relative Modality.}}
    \label{fig:CBP-payment-request-parent-modality}
\end{figure*}

\chadded[id=add]{Fig~\ref{fig:CBP-payment-request-parent-modality} depicts the CBP results when running the \texttt{relative} modality on the same variants.
Normalizing activity timestamps on the immediate parent, based on the process model, effectively ``cuts off'' chains of three or more dependencies, revealing causal relations only between any two directly-followed activities. This eliminates any collider or confounder pattern from the results.}

%% file: cpm4sax-related-work.tex
\section{Related Work}
\label{sec:related}

Our novelty resides in applying \chdeleted[id=del]{causal discovery}\chadded[id=add]{CD} techniques in order to find causal execution dependencies in event logs, so our research is positioned at the intersection between PD and CD. While to the best of our knowledge, there is no related work in this specific space, we bring some notable works that associate \chdeleted[id=del]{causal discovery}\chadded[id=add]{CD} and causal inference with business process management.
Most of the PD approaches apply some threshold over the time precedence (counting over ‘directly follows’ or ‘eventually follows’ relationships). ``Causal'' relations \chadded[id=add]{are}\chdeleted[id=del]{is} sometimes used to express the frequencies of these time-precedence dependencies. One such example is \cite{Kourani2023} where there is a reduction in the representational bias of the hybrid miner algorithm by exploiting causal graph metrics to mine for long-term dependencies. We suggest following a different approach that is based on a bidirectional (asymmetric) observation of the relationship between the time property of any two activities.
It is important to note that our core meaning of the causal relationships is different. The novelty of our work resides in applying \chdeleted[id=del]{causal discovery}\chadded[id=add]{CD} techniques (e.g., LiNGAM) to determine the causal execution dependency between any two activities only by observing a series of task execution times.

Causal-nets~\cite{van2016process,cnets2011}, introduced in 2002 by van de\chadded[id=add]{r}\chdeleted[id=del]{t} Aalst, are a form of process representation (i.e., a notation) that is expressively richer than a variety of modeling languages (e.g., Petri-nets, BPMN, and EPC), 
and could serve as a viable alternative representation of CBPs. 
However, being merely a notation, causal-nets do not include any mechanism to infer the causal structure among activity executions, and the knowledge about the causal structure should be either provided by domain experts or by using some discovery algorithm as tackled in our work\chadded[id=add]{.}

As mentioned in the background section, methods for causal inferencing and discovery are split between techniques that are used to assess the quantitative extent of the impact of an intervention and techniques to determine if qualitative causal relationships exist with respect to a given dataset with no further intervention. A technique such as the Conditional Average Treatment Effect (CATE)~\cite{cate2019} 
gives a measurement to assess the magnitude of an intervention. 
For example, the work in~\cite{bozorgi2020process} used CATE to create an uplifting tree that maximizes the CATE in the splits, to determine the best action to be taken. 
The work assumed that the confounding set is given \chdeleted[id=del]{(i.e., not spurious)} while we discover it using LiNGAM. 
In ~\cite{Shoush2022,Shoush2023} CATE is used for prescriptive process monitoring, to prevent interventions from being triggered unnecessarily when the level of confidence is insufficient. Here as well, the causal model is given, while we discover it applying LiNGAM on the execution times of the activities in the process. 

In \cite{narendra2019counterfactual}, the authors used causality to answer what-if questions for process improvement relying on a given process model, 
\chdeleted[id=del]{where}\chadded[id=add]{while} we focus on its discovery out of the event logs. 
In~\cite{hompes2017}, the authors propose a technique that generates a graph of causal factors explaining process performance. To detect causal relations, they test for Granger causality, a statistical test widely used for causal analysis of time series. The idea is that values of performance indicators are seen as time series while 
we use \chdeleted[id=del]{causal discovery}\chadded[id=add]{CD} to determine dependencies among task execution times. \chadded[id=add]{In other words, in~\cite{hompes2017} the relations between variables are determined based on correspondences among timestamped values (of the KPIs) rather than solely between the timestamps (of the tasks) themselves as in our method}. 

In \cite{Effendi2019} and \cite{Effendi2020} an alternative way to discover a business process model from a double timestamp event log (i.e., \chadded[id=add]{in which} there is a start time and complete time of each activity) is proposed. The algorithm is based on ``temporal causal relations'', which \chdeleted[id=del]{is a pattern of event log that occurs from activities performed in the process.}
\chadded[id=add]{are flow patterns reflective of different flow structures based on types of time-precedence relations among activities in the event log (e.g., same start time, same end time, contains, overlaps).}
These patterns can reveal sequential and parallel AND, OR, and XOR relations. 
\chdeleted[id=del]{This differs from our approach which applies \chdeleted[id=del]{causal discovery}\chadded[id=add]{CD} techniques (e.g., LiNGAM) to discover the causal process view to highlight causal and non-causal relations in the mined process model.}
\chadded[id=add]{This work uses a modified version of the Alpha Miner called Modified Time-based Alpha Miner that incorporates timing information to better handle real-world complexities in process mining. As with the core comparisons presented in our synthetic examples, this approach still retains the fundamental difference stemming from a process model constructed based on the frequency and order of activities, not on establishing a causal statistical relationship between the timing of those activities. Thus, even with timestamps, LiNGAM would not inherently construct process models or reveal the kind of direct-follows relationships typical in process mining but would rather attempt to infer causal directions by utilizing the statistical properties of the timing, specifically assuming linear dependencies and non-Gaussian distribution of the errors, which is conceptually and methodologically distinct from the goals of process mining.}

Another work that proposes an alternative way to process mining is \cite{KOORN2022}. Here, the authors propose the ARE algorithm for process mining that identifies action-response-effect patterns in order to understand relevant aspects of process execution beyond the control flow perspective. To achieve this, they rely on statistical relations discovered from the event logs through some statistical tests \chadded[id=add]{stemming from time adjacency between the events as a signal for cause-effect dependence}. While the ARE algorithm is \chadded[id=add]{essentially a frequency-based discovery of the directed follows relations with cause-effect pairs already given in the log, our method relies merely on the functional (linear) relation among the event timestamps without necessarily assuming any prior knowledge about the relationship among the tasks.} \chdeleted[id=del]{based on statistical tests and aims at discovering the causes of certain events, it is not based on causal execution dependencies discovered among the activities in the data.}

\chdeleted[id=del]{\cite{ho2023inferring} propose\chadded[id=add]{s} a modeling approach for event datasets where events have detailed temporal information, including occurrence time and duration. Using this model, the authors can describe causal relations between events where both cause and effect span a finite period of time. While the approach does not address process activities, but rather events that are marked by two timestamps, it can provide an alternative for the LiNGAM algorithm applied by us.}

In \cite{Waibel2021}, causal event models are introduced to correctly capture the $n:m$ relations among events stored in a relational database. Causal relations are identified by the use of foreign key relationships. More recently, the work in \cite{waibel2022causal} employs relational databases to tackle the problem of spurious relations.
It demonstrates that directly-follows miners produce numerous spurious relationships that can be reduced. Our approach differs in some aspects\chdeleted[id=del]{:}\chadded[id=add]{.} First, our notion of causality is based on causal execution dependencies and not \chadded[id=add]{on} time precedence. Second, we base our \chdeleted[id=del]{causal discovery}\chadded[id=add]{CD} solely on the input event log. Third, our aim is to produce a new view that emphasizes the causal execution dependencies rather than altering the process model discovery itself.

With respect to PD algorithms, we employed $\alpha$ and Heuristic \chadded[id=addr2]{on the synthetic data}. \chadded[id=addr2]{We acknowledge} there are \chdeleted[id=del]{many extensions to these}\chadded[id=add]{more contemporary} algorithms, such as Flexible Heuristics Miner (FHM)~\cite{fhm2011}\chadded[id=add]{, Inductive Miner~\cite{Leemans2017}, and Split Miner~\cite{Augusto2019}}, being more robust to \chdeleted[id=del]{noise in the} \chadded[id=add]{noisy and incomplete} data. For the sake of establishing the core hypothesis, the two algorithms selected are representative, while for the Sepsis analysis, we employed a commercial tool.

\chdeleted[id=del]{Causal-nets[] are a form of process representation (i.e., a notation) that is expressively richer than a variety of modeling languages (e.g., Petri-nets, BPMN, and EPC),}
\chdeleted[id=del]{and could serve as a viable alternative representation of CBPs.} 
\chdeleted[id=del]{However, regardless of the notation employed, the knowledge about the causal structure should be either provided by domain experts or by using some discovery algorithm as tackled in our work.}

While we used the LiNGAM algorithm for \chdeleted[id=del]{causal discovery}\chadded[id=add]{CD}, 
there are other alternatives recently developed and applied in the context of causal analysis in business processes. For example, the work in~\cite{root2020} shows how explanations (about effects of changes) could be facilitated by causal structures describing a variety of features about the process execution. 
Concretely, this work demonstrated the viability of the approach using a Greedy Fast Causal Inference (GFCI)~\cite{GFCI} algorithm\chadded[id=add]{.}\chdeleted[id=del]{ that u}\chadded[id=add]{U}nder \chadded[id=add]{the assumptions of} a large sample \chdeleted[id=del]{limit}\chadded[id=add]{size,} \chdeleted[id=del]{and conforming to} linear dependencies among features\chadded[id=add]{,} and additive noise\chadded[id=add]{, this algorithm} can yield faithful results.  

Another example is the work in~\cite{caise2022} that looks into inferring causal relationships among decision points (i.e., splits) in the process
that aims to determine causal dependency chains along process decisions to set the probability a decision could \chdeleted[id=del]{have on}\chadded[id=add]{affect} process outcome, employing the Missing Value PC (MVPC) \chdeleted[id=del]{causal discovery}\chadded[id=add]{CD} algorithm~\cite{pmlr2019}, an extension to PC~\cite{spirtes2000causation} that starts from a fully connected undirected graph over all variables, and gradually removes edges based on causal independence statistical tests. 
\chdeleted[id=del]{While PC may yield equivalent classes of possible DAGs, 
the LiNGAM algorithm can further conclude a unique DAG for a given dataset.}
The main underlying assumption in~\cite{caise2022}, is that the process model is given and is correct while we apply process discovery techniques to mine the process model and apply \chdeleted[id=del]{causal discovery}\chadded[id=add]{CD} techniques to mine
the causal process model out of the same event log and use the latter as a reference to test whether any relation in the discovered/mined process model is causal or not from \chadded[id=add]{an} execution time perspective. 

\chadded[id=add]{Both MVPC and GFCI algorithms belong to a family of constraint-based algorithms, yielding a set of causal graphs that are consistent with the data in the form of a Partial Ancestral Graph (PAG). 
While both algorithms may yield equivalent classes of possible DAGs, 
the LiNGAM algorithm infers a unique DAG for a given dataset, making it more appropriate for the purpose of discrepancy analysis with the BP.}

%% file: cpm4sax-conclusions.tex
\section{Conclusions and Future Work}
\label{sec:conclusions}




Traditionally, process discovery techniques have been associational rather than causal, limiting reasoning about potential process changes thus affecting process improvement capabilities.

The ability to reason about the causes leading to certain consequences in a business process largely depends on the ability to infer the correct chain of event executions that affect these consequences. 
Current discovery techniques use time precedence as the primary guiding rationale for determining task ordering in the discovered process model. In this paper, we show that relying merely on time precedence can sometimes deviate from the real causal dependencies in the business process model, and as a result, to inadequate execution interpretations.
We provide a method to reveal the causal execution dependencies present in an event log by applying the LiNGAM algorithm.
Respectively, we also provide a method for constructing a new view that overlays the discovered process model with causal relationships, which is agnostic to the concrete PD algorithm employed.

To the best of our knowledge, our work is a first attempt at applying state-of-the-art \chdeleted[id=del]{causal discovery}\chadded[id=add]{CD} algorithms to the timing of activities in order to detect, and respectively highlight, causal execution dependencies among tasks over the output of process discovery techniques.

We envision different directions as future work. First, 
we could relax the assumption of known common causes (e.g., observed confounders) to test for latent causal tasks. 

Second, this work 
calls for an empirical user study to test the interpretability and usefulness of the proposed discrepancy view.

Third, \chadded[id=add]{one} may leverage CBPs for the sake of process outcome explainability. We have previously shown~\cite{amit2022} that XAI frameworks such as LIME can be extended for better business process explainability. A similar approach was also employed to SHAP~\cite{heskes2020causal}. We could improve this na\"ive approach 
that bundled all attributes in the input vector to the explainer to stratify the attributes 
based on the true causal dependencies as discovered in the underlying causal model. 


Last, the discovery of the actual causal process execution can yield a significant means to drive more logically compact explanations (i.e., sufficient reasons~\cite{rubin2022}). 
The imposing of causal dependency constraints over the overall space of explanations\chdeleted[id=del]{,} 
can help remove explanations that are logically subsumed by other explanations that are shorter, do not embed redundant reasons, and as a result, are also likely to be more comprehensible.

Causal execution dependencies play a vital role as an instrument for a deeper understanding of process executions, and a vehicle for process improvement. 
Our ultimate goal is to enrich current PM tools with a CBP-view overlaying feature.






